%% file: ARXIV_BENCH.tex
\documentclass[journal]{IEEEtran}
\usepackage{subfig}
\usepackage{url}
\usepackage{nicefrac}
\usepackage{graphicx}
\usepackage{tabularx}
\usepackage{afterpage}

\usepackage{amsmath,amsfonts,amsthm,amssymb} % Math packages
\usepackage{rotating}
\usepackage{fancyhdr}  
\usepackage{tikz}
\usepackage{mathtools}
\usepackage{cite}
\usepackage{pgfplots}
\usepackage{algorithm}
\usepackage[noend]{algpseudocode}
\usepackage{siunitx}

\newcommand*\mean[1]{\overline{#1}}
\newtheorem{defn}{Definition}[section]

\makeatletter
\def\BState{\State\hskip-\ALG@thistlm}
\makeatother

\pgfplotscreateplotcyclelist{mycolorlist}{
red,
blue,
teal,
orange,
black
}

\usepackage{xcolor}
\newcommand \blue[1]{\textcolor{black}{#1}}
\newcommand \red[1]{\textcolor{black}{#1}}

\definecolor{gaussianLocError_1}{HTML}{9ecae1}
\definecolor{gaussianLocError_2}{HTML}{3182bd}

\newcommand\gauss[2]{1/(#2*sqrt(2*pi))*exp(-((x-#1)^2)/(2*#2^2))}
\title{\Large \bf 
Predicting Performance of SLAM Algorithms}

\author{Matteo Luperto $^{1}$, Valerio Castelli$^{2}$, and Francesco Amigoni$^{2}$% <-this % stops a space
\thanks{$^{1}$M. Luperto is with the Applied Intelligent Systems Laboratory, Universit\`{a} degli Studi di Milano, Via Festa del Perdono 7, 20122 Milano, Italy 
	{\tt\small matteo.luperto@unimi.it}}
\thanks{$^{2}$F. Amigoni and V. Castelli are with the Artificial Intelligence and Robotics Laboratory, Politecnico di Milano, Piazza Leonardo da Vinci 32, 20133 Milano, Italy
        {\tt\small francesco.amigoni@polimi.it},
        {\tt\small valerio.castelli@mail.polimi.it}}
}%
\begin{document}

\maketitle
\thispagestyle{empty}
\pagestyle{empty}

%%%%%%%%%%%%%%%%%%%%%%%%%%%%%%%%%%%%%%%%%%%%%%%%%%%%%%%%%%%%%%%%%%%%%%%%%%%%%%%%
\begin{abstract}
Among the abilities that autonomous mobile robots should exhibit, map building and localization are definitely recognized as fundamental. Consequently, countless algorithms for solving the Simultaneous Localization And Mapping (SLAM) problem have been proposed. Currently, their evaluation is performed \emph{ex post}, according to outcomes obtained when running the algorithms on data collected by robots in real or simulated environments. 
In this paper, we present a novel method that allows the \emph{ex ante} prediction of the performance of a SLAM algorithm in an unseen environment, before it is actually run. Our method collects the performance of a SLAM algorithm in a number of simulated environments, builds a model that represents the relationship between the observed performance and some geometrical features of the environments, and exploits such model to predict the performance of the algorithm in an unseen environment starting from its features. 
\end{abstract}

%%%%%%%%%%%%%%%%%%%%%%%%%%%%%%%%%%%%%%%%%%%%%%%%%%%%%%%%%%%%%%%%%%%%%%%%%%%%%%%%

\input{01-Intro}

\input{02-Art}

\input{03-Our}

\input{04-Data}

\input{05-Method}

\input{06-Exp}

\input{07-Con}

%\input{newimg}


%%%%%%%%%%%%%%%%%%%%%%%%%%%%%%%%%%%%%%%%%%%%%%%%%%%%%%%%%%%%%%%%%%%%%%%%%%%%%%%%

\bibliographystyle{IEEEtran}
\bibliography{citations2}

%%%%%%%%%%%%%%%%%%%%%%%%%%%%%%%%%%%%%%%%%%%%%%%%%%%%%%%%%%%%%%%%%%%%%%%%%%%%%%%%

%%%%%%%%%%%%%%%%%%%%%%%%%%%%%%%%%%%%%%%%%%%%%%%%%%%%%%

\end{document}

%% file: 01-Intro.tex
\section{Introduction}\label{sec:INT}

\IEEEPARstart{M}{ap} building and localization are two of the most fundamental abilities required for autonomous mobile robots. They are usually addressed together by solving the Simultaneous Localization And Mapping (SLAM) problem, which has been extensively studied in the last decades~\cite{Thrun:2005:PR:1121596}. However, the availability of countless SLAM algorithms contrasts with their relatively less mature assessment, evaluation, and benchmarking, which are topics currently of high interest in robotics~\cite{fabio-ram}. The state-of-the-art methods for evaluating SLAM algorithms (e.g.,~\cite{Balaguer_towardsquantitative,Kuemmerle2009,Schwertfeger2016}) perform \emph{ex-post} assessments of their performance, namely they run a SLAM algorithm on data collected by robots in simulated or real environments and evaluate the obtained results according to some performance metric, often involving comparison with ground truth. 

In this paper, we present a novel method that makes a step forward and provides the expected performance of a SLAM algorithm in new environments, before it is actually applied to data coming from those environments. In this sense, the proposed method opens the possibility to perform a \emph{predictive benchmarking} of SLAM algorithms. We call predictive benchmarking the capability to anticipate the performance results obtained by a system on a benchmark, in our case by a SLAM algorithm (on a given robot platform) in a previously unseen environment. The main original contribution of this paper is thus a method that uses the results of a SLAM algorithm in several environments to predict its performance in new ones. 
 
\red{Compared to current SLAM evaluation techniques, predictive benchmarking has the advantage of estimating the expected performance of SLAM algorithms in new environments without requiring their actual exploration.}

The proposed method works by collecting the performance of a SLAM algorithm on a number of environments, by building a model that represents the relationship between the measured performance values and the features of the environments, and by exploiting such a model to predict the performance of the algorithm starting from the features of an unseen environment. 
In particular, we consider well-known \red{2D SLAM algorithms} performing simulated runs on a wide range of different environments to obtain the training data, and discover that simple linear models can capture the relationship between the skeleton of an environment, represented by its Voronoi graph, and the quality of its map returned by SLAM algorithms. We evaluate our method on two widely adopted SLAM algorithms, GMapping~\cite{Grisetti:2007:ITG:2211651.2211962} and Karto~\cite{KARTO}, and we validate our method on data from publicly available datasets and collected with real robots.

Two are the main benefits of the proposed method for predictive benchmarking, which also represent the motivations for developing it. First, the availability of reliable predictions on the performance of SLAM algorithms can help developers of robot systems to make informed decisions at design time, reducing expensive trial-and-error approaches. Second, knowledge of environments in which a SLAM algorithm is expected to perform well and in which it is expected to perform badly can be used to establish the boundaries of its applicability. This can enable the generalization of results~\cite{Amigoni:2017:TGE:3065975.3066115}, which currently is a rather limited practice in many areas of robotics, hindering the possibility of extending the results obtained in a setting to other settings. \red{The use of methods for predictive benchmarking can be adapted to assess the choice of sensors, robot configuration, and algorithms at the design phase of the robotic system, thus having a preliminary evaluation of performances event at a pre-prototypal phase of a newly developed platform. Finally, fact that our method exploits linear models enables the interpretability of the results, thus increasing their applicability to real-world settings.}

The paper is structured as follows. The next section surveys the relevant related work on the evaluation of SLAM algorithms.  Section~\ref{sec:OUR} overviews the proposed method, which is detailed in Section~\ref{sec:DATA}. The experimental validation of the method is reported in Section~\ref{sec:EXP}. Finally, Section~\ref{sec:CON} concludes the paper.
\red{A preliminary version of this work has been presented in \cite{OURIROS}, where is described the methodology adopted here to collect training data and to benchmark performances of SLAM algorithms.
}

%% file: 02-Art.tex
\section{Related Work}\label{sec:ART}

As the number of algorithms developed to solve the SLAM problem keeps growing, finding reliable ways to assess their performance has become a significant challenge for the robotics research community. This problem is complex, as different SLAM algorithms applied to the same data may produce different results in terms of map format and quality, as shown in \cite{6719348}. Ignoring for the moment how data are acquired by robots, two main aspects are involved in evaluating SLAM algorithms.
 
The first aspect is about the environments in which tests are conducted. One approach is to perform robotic competitions, where a custom evaluation environment is created to either mimic a real scenario or be particularly hard for robots to handle. The robots are then assigned tasks which fulfillment degree can be used for indirectly evaluating the performance of localization and mapping, as happened for example in the RoboCup@Home competition~\cite{robocup-home-ai}. \red{As the quality of the reconstructed map affects the ability of the robot to successfully complete the task, it is assumed that the systems that achieve the highest overall performance are also the best performing on the SLAM subtask. This approach has been used to evaluate, among others, the performance of cleaning robots [IROS, 2002], self-driving cars in an urban area [Darpa, 2007], rovers moving in a simulated Mars setting [ESA, 2008], and robots operating in Urban Search and Rescue scenarios [RoboCup Federation, 2009].}

A competition that separates the evaluation of task performance from that of localization performance of a robot has been proposed in RoCKIn~\cite{rockin-ram}.
\red{Two important shortcomings of robotic competitions are their limited scale and little resemblance to real application environment (most competitions being held in relatively small laboratories) and is the significant complexity and diversity of hardware and software settings that have an impact on the robots' performances (as each robot has a different hardware architecture, software stack, and selection of navigation and mapping parameters). For these reasons, it is extremely difficult to say how much a robot's performance is due to the choice of the SLAM algorithm and how much is due to other factors (that may depend on that particular execution of the task, and may not be consistent wrt other repetitions) instead.}

Another solution adopted by the robotics community is the collection and publication of datasets to be used as benchmarks. The use of standardized environments enhances reproducibility, replicability, and comparability of the results~\cite{paper-con-monica}. Examples include Radish~\cite{Radish}, Rawseeds~\cite{Ceriani2009},  
datasets from the Computer Vision Group of the TUM~\cite{sturm12iros}, and openSLAM\footnote{\url{https://openslam.org}}.
\red{However, most of those datasets available do not include ground truth data (like the followed trajectory and the floor plans of the buildings), effectively preventing the evaluation of many performance metrics that are based on some form of a priori knowledge about the actual map and trajectory. Moreover, commonly available datasets may suffer of generalizability issues, which requires the identification of datasets with the same characteristics as those of the target building where the robot will be used. Datasets are usually collected into controlled (small) environments, like specially fitted laboratory rooms; this limits the scale and diversity of the tests that can be reasonably conducted with such data, which fail to replicate the complexity of many real-world application environments. }

%\emph{Simulation} has been proposed as a possible solution to this problem, leading to the development of several robotic simulators. 
A third approach, which allows tests in arbitrarily complex environments is the use of simulations. 

Some of the most popular simulation tools are Stage\footnote{\url{http://wiki.ros.org/stage}} (in 2D), \red{a lightweight robot simulator that can handle populations of hundreds of virtual robots in a two-dimensional bitmapped environment,} 
%allows to simulate the effects of a uniformly distributed random error on odometry readings, 
%The suite can also be used in conjunction with ROS. 
%To overcome the limitations of two-dimensional simulations, Howard et al. introduced 
Gazebo\footnote{\url{http://gazebosim.org}} (3D), \red{ an open-source robotic simulator that recreates 3D dynamic multi-robot environments that accurately mimic real-world physics and offer the possibility to simulate a broader range of actuators and sensors}, and
USARSim (3D) \cite{4209284}, an alternative robotic simulator developed to support the virtual robots competition within the RoboCup initiative.
\red{More recently, the use of 3D realistic simulated environments obtained from real-world 3D models has been used in several applications, particularly those related to tasks that have to be performed in 3D (photo-)realistic environments. Notable examples of such datasets are those based on the Mattertport or on the SUNCG dataset\cite{chang2017matterport3d,wu2018building,savva2017minos,habitat19iccv}. However, those simulated environments are particularly useful for assessing robots performance that rely mostly on vision (as Embodied Question Answering, EQA) but do not present a realistic method for simulating other robot characteristics as odometry or 2D laser range scanners.}

The second aspect is about the performance metrics employed to evaluate SLAM algorithms. Several approaches have been proposed over the years. According to \cite{4209739}, in order to effectively evaluate and compare different SLAM algorithms it is necessary to: (i) provide extensive information about the produced maps, (ii) report the behavior of the mapping system for different values of the parameters, (iii) include one or more examples of maps produced following a closed-loop path, and (iv) whenever a ground truth map is available, use it to assess the quality of the produced map by evaluating its distance from the ground truth map. 

In \cite{Balaguer_towardsquantitative}, the performance of three SLAM algorithms is assessed by visually estimating the fidelity of the reconstructed maps with respect to the ground truth.

In \cite{4433772}, it is argued that comparing the map produced by a SLAM algorithm to its ground truth counterpart is not an appropriate evaluation metric, as the main purpose of SLAM algorithms is not to produce human-understandable maps. 
\red{In this context, the degree of accuracy of a produced map with respect to a known ground truth does not necessarily reflect its usefulness, as schematic representations of the environment could still be sufficient for the completion of the desired task while more visually faithful depictions could lack details, like doors and passages, that are crucial for the successful completion of the task.}
The authors of~\cite{4433772} propose to evaluate the usefulness of maps by measuring the degree to which paths created in the generated maps can be actually traversed in the real world \red{by using two different metrics: the first is the degree to which invalid paths created in the generated map would cause the robot to collide with a structural obstacle in the real world; the second is the degree to which the robot should be able to plan a path from one position to another using the generated map, but cannot because such paths are invalid in the ideal map. However, this approach also suffers from two main limitations, as it can only be applied to SLAM algorithms that produce occupancy grid maps as outputs and it requires precise alignment of the produced map with the ground truth map.} 

In \cite{Fontana2014}, it is proposed to adopt context-dependent  performance metrics that capture the ability of SLAM algorithms to produce useful maps for localization and navigation. The authors define a self-localization error and an integral trajectory error, which both require ground truth trajectory information. 

%A significant drawback relative to metrics that rely on absolute poses is that they are strongly influenced by the timestamp at which an error occurs. This introduces a strong bias in the evaluation, as an error of a given entity may lead to very different results depending on whether it is introduced at the beginning, in the middle, or at the end of an exploration. Consider for instance a rotation error of a few degrees: if the error is introduced at the end of an exploration, its effect will be limited to a very small number of poses and its impact on the overall performance will be negligible; if however it occurs at the beginning of the exploration, the reconstructed trajectory will rapidly diverge from the ground truth one, leading to a much higher measured error despite the map still being substantially correct.  

In \cite{Kuemmerle2009}, a metric that considers the deformation energy that is needed to superimpose the estimated trajectory onto the ground truth trajectory, i.e., that is based on the relative displacements between poses, is proposed. We detail it in Section \ref{sec:DATA}, since we use an extended version of this metric. \red{However, Kuemmerle et al. do not provide a definite criterion to choose which relative displacements should be considered to compute the metric, noting that, in absence of ground truth information, close-to-true relative displacements can be obtained by other sources of information, such as background human knowledge about the length of a corridor or the shape of a room.} 

In the context of SLAM evaluation through map evaluation, \cite{Schwertfeger2016} investigates the possibility to use topology graphs derived from Voronoi diagrams to capture high-level spatial structures of indoor environments. 
%A Voronoi diagram is a partition of the space into \emph{cells}; each cell contains all points of the map whose distance to a target obstacle is not greater than their distance to all other obstacles. The graph is then obtained by considering the boundaries of said cells. 
In \cite{6719348}, the performance of five SLAM algorithms are evaluated \red{- HectorSLAM, Gmapping, KartoSLAM, CoreSLAM and LagoSLAM -}  in terms of distance between the generated map and the ground truth map using a metric based on k-nearest neighbor.  
 
\red{Finally, \cite{DBLP:journals/corr/abs-1708-02354} proposes to evaluate the quality of generated maps by measuring three different aspects: the proportion of occupied and free cells to determine blur, the amount of corners in the map, and the amount of enclosed areas. }

%\textcolor{red}{Citare lavoro di tesi di quello di Stanford che parla di SLAM evaluation e dice che EX-post evaluation è un problema. vedere se ci sono suoi paper a riguardo?}
%As documented in the previous section, the research community has not agreed on a single definition of SLAM algorithm performance, proposing instead several metrics to measure different aspects of SLAM algorithms behavior and a number of possible ways to assess them. However, 
A trait common to all the above solutions to SLAM evaluation is that they work on already collected data, i.e., they assess the performance of a SLAM algorithm in a certain environment only after the environment has been actually visited. This approach, although valid, presents some limitations.
\emph{Ex post} evaluations typically need to gather both sensory and ground truth data or require some manual intervention in the evaluation process\red{, as obtaining ground truth trajectory data requires an accurate tracking system pervasive in the environment in order to be measured with sub-millimeter precision.}% An example of such a system is \emph{OptiTrack}\footnote{\url{https://www.optitrack.com}}, a positioning system developed by NaturalPoint, Inc.\footnote{\url{https://www.naturalpoint.com}} in 2009 that uses a variable number of synchronized infrared cameras, each containing a grayscale CMOS imager capturing up to 100 FPS, to triangulate the position of an infrared reflector placed on the robot itself. 
\red{As the environment becomes larger, the number of required cameras to obtain such ground truth measures grows both expensive and complex to operate. Moreover, as we show, the performance of a SLAM algorithm in an environment may in fact vary across multiple explorations, so that an individual exploration of the environment may not accurately represent the actual average performance of the algorithm in that setting and repeated measurements require to set up the system at every run.}%% and can be often executed in parallel on dedicated servers to increase the throughput, while still providing reasonably accurate results \cite{Balaguer_towardsquantitative,Jaulmes_quantitative,7822338}. However, there is a third fundamental limitation of these evaluation methods that affects both real-world and simulated explorations: 
%Another limitation of ex post evaluation methods is their lack of generalization capabilities. 
Generalization of results, intended as the identification of significant correlations between the performance of a given SLAM algorithm in different scenarios, is also difficult~\cite{Amigoni:2017:TGE:3065975.3066115}. 
%A \emph{setting} is defined by the environment in which the algorithm is tested, the type the robot sensors, and the specific values of the parameters. Each of the examined evaluation methods proposes a different and potentially equally valid way to assess the a posteriori performance of a SLAM algorithm after an exploration has taken place; however, 
None of the above evaluation methods provides a way to use previously collected performance data to make predictions about the performance of a SLAM algorithm in a scenario before an exploration actually takes place.
\red{This limitation is significant, since knowing how well a SLAM algorithm performs in a certain setting does not immediately provide any information as to how well it will perform in a different setting, as the measured level of performance is strictly dependent on the characteristics of the setting itself. 
The problem is exacerbated by the fact that there is no single best SLAM algorithm for every scenario. As an example, a study \cite{7822338} conducted by Turnage using the Hausdorff Distance between the ground truth map and the reconstructed map as a metric to assess the relative level of performance of HectorSLAM, CoreSLAM and Gmapping on three different environments shows that while HectorSLAM outperforms both CoreSLAM and Gmapping in two environments out of three, CoreSLAM performs best in the remaining one, with HectorSLAM and Gmapping performing the same.}
This paper contributes a method to perform such a prior assessment of the performance of a SLAM algorithm. 
%is crucial towards enabling more pervasive applications of mobile robotics, as it vastly simplifies the deployment of robots in real-world contexts. The absence of generalization in SLAM performance assessment increases the difficulty in knowing at design time which SLAM algorithm and robot configuration best fits a given application scenario, often requiring a cumbersome and expensive trial and error process.

In the literature, some efforts has been made to predict robot performance in previously unseen scenarios. %\textcolor{red}{citare paper path trasversal da aerial image del giusti?}
%Research has been done on the broader topic of promoting generalization in robotics. On one hand, there have been attempts to explicitly improve the generalization capabilities of algorithms for robotic applications. 
%An example is \cite{Pinville:2011:PGE:2001576.2001612}, where is proposed a supervised learning approach to improve the generalization capabilities of controllers in Evolutionary Robotics. 
%On the other hand, researchers have focused on the problem of developing approaches to predict robotic performance in a wide variety of contexts. 
%\red{In \cite{Chandrasekaran2010}, the authors investigate the usage of neural networks, fuzzy systems, genetic algorithms, and other soft computing techniques to predict the performance of several industrial machining processes.}
An example is \cite{7484706}, where a model to assess the traversal cost of a natural outdoor environment for an autonomous vehicle using A* planning is proposed. 
\red{The model exploits information about the complexity of the environment itself, including slope and the presence of vegetation, to ultimately predict the average speed of the vehicle.}
Considering autonomous wheeled robots, in \cite{8098664} the authors introduce a method to estimate the traversal time of a path by using its length, smoothness, and clearance as features of a non-linear regressor. 
To the best of our knowledge, there is currently no method that predicts the performance of SLAM algorithms, as we do in this paper.
%The model, which is trained by using Gazebo simulations of an omnidirectional robot in a variety of maps, is shown to significantly outperform predictions only based on path length. %In \cite{dawson_prediction}, Dawson et al. highlight the limits of several performance prediction systems in estimating the average coverage time of an environment in multi-robot autonomous exploration, citing the difficulty for simulations to properly take into account the slowdowns introduced by inter-robot communications, physical interferences and network latency. 
%In \cite{Amigoni:2017:TGE:3065975.3066115}, authors argue that a major hurdle towards achieving a higher level of generalization of experimental results in autonomous robotics is the limited representativeness of the experimental settings. This is especially true for physical experiments involving actual robots, which are often conducted in small controlled settings that are little representative of the complexity encountered in real-world applications.

%% file: 03-Our.tex
\section{Our Method for Predictive Benchmarking}\label{sec:OUR}

%SPIEGONE - CAPPELLO INTRODUTTIVO. DA SCRIVERE. COSA DIRE.

%per risolvere il problema del predictive benchmarking il nostro metodo esegue una serie di step sequenziali, che sono visibili in figura \ref{fig:model_creation_bpmn}. gli step sono: otteninmento di una serie di dati sul GT e di metriche di valutazione di slam; estrazioni di feature da delle planimetrie. creazione di un modello che colleghi metriche di valutazione di SLAM alle feature dell'ambiente;
%validazione e testing.
%Queste tre cose sono in Section 4, Section 5 e Section 6 (rispettivamente). Copioincollo una parte che possiamo usare come fonte per scrivere questa sezione.

\begin{figure}[t!]
\centering
\includegraphics[scale=0.3]{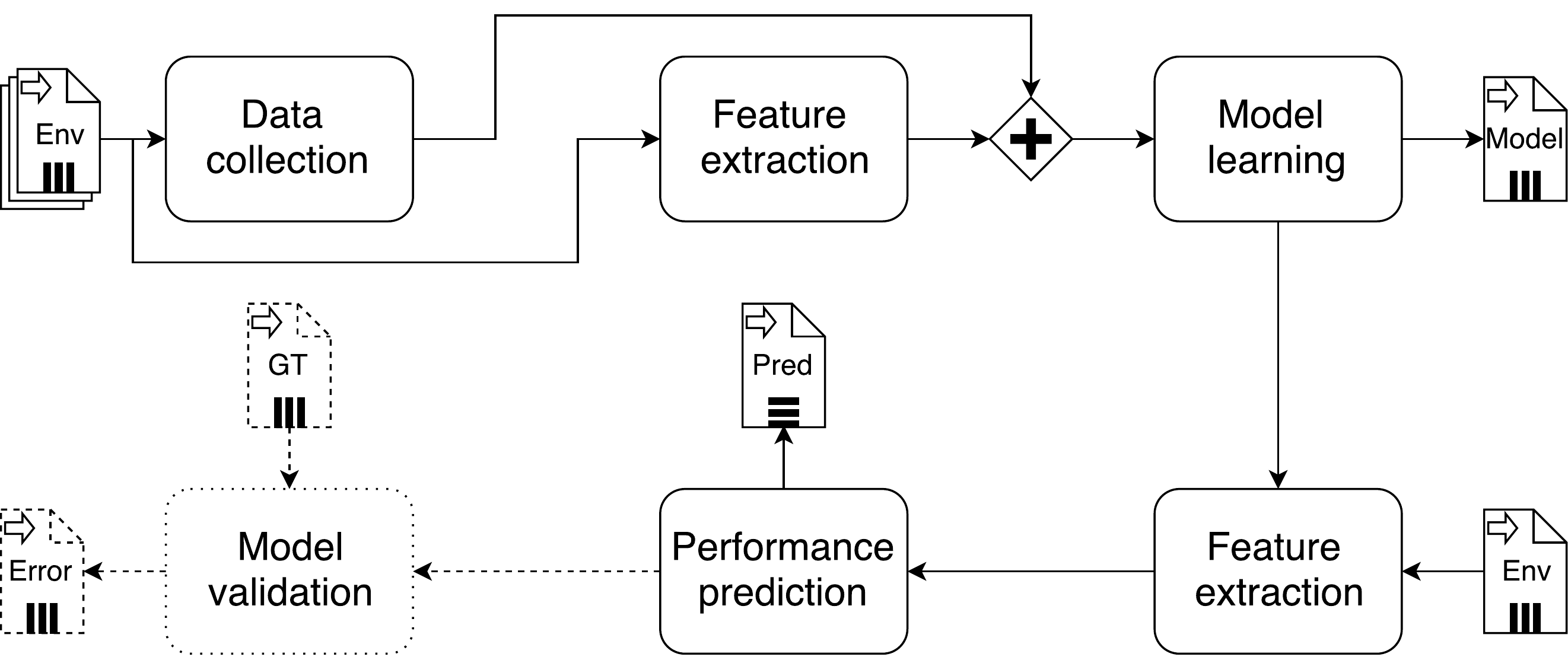}
\caption{The conceptual flow of our method for predictive benchmarking. Dashed components are optional.}
\label{fig:model_creation_bpmn}
\end{figure}

%The goal of this paper is to make a step beyond the state of the art by developing a method that predicts the performance of SLAM algorithms in unvisited environments.
Our predictive benchmarking method uses the performance results obtained by a SLAM algorithm in a number of environments to predict its performance in new ones. 
\red{Compared to current SLAM evaluation techniques, which are used to perform ex-post benchmarking, our \emph{predictive benchmarking} approach provides an estimate of the expected level of performance of a SLAM algorithm in new environments without requiring their actual mapping. 
While existing SLAM evaluation techniques handle every new environment as independent, predictive benchmarking builds upon the knowledge gained with previous evaluations to create a model of the relationship between environments and their associated SLAM performance; this results in increasingly accurate performance predictions as the number of explorations used for training increases.} 
\red{It consists of a sequence of modules, each responsible for a logically distinct step, as shown in Fig. \ref{fig:model_creation_bpmn}. In this paper, we consider 2D lidar-based SLAM methods, using GMapping~\cite{Grisetti:2007:ITG:2211651.2211962} as the reference algorithm  because of its widespread use and the availability of stable implementations. Validation of the method is performed by considering another widespread SLAM method, KartoSLAM}\footnote{\url{http://wiki.ros.org/slam_karto}} \blue{\cite{KARTO}, but the proposed approach potentially works with any other SLAM algorithm. The choice of these two SLAM algorithms is motivated by the fact that those methods are widely used both for research purposes and for real-world applications \cite{STRANDS},\cite{10.1007/978-3-319-97589-4_24}. }
%\textcolor{red}{CAMBIARE: dire che usiamo prevalentemente gmapping come riferimento, ma che usiamo quello e KartoSLAM in parallelo per provare la generalizzabilità. Non li metterei sullo stesso piano, terrei comunque i due leggermente separati. Il discorso poi diventa così generale: dove parliamo di GMAPPING passiamo a SLAM algorithm. DIrei invece che ci concentriamo su ambiente indoor 2D}
%QUESTO PARAGRAFO E' UNA MOTIVAZIONE E ANDREBBE SPOSTATO NELL'INTRODUZIONE. This approach significantly simplifies the development and deployment of autonomous mobile robots by allowing designers and manufacturers to evaluate at design time the suitability of a certain choice of SLAM algorithm, parameters values and sensors for a given scenario. It may also be used as a basis to perform comparisons between SLAM algorithms at a lower cost and on a broader set of environments than what is possible with traditional SLAM evaluation techniques. Finally, it can be used to estimate the localization accuracy of a SLAM algorithm in real-world environments without requiring ground truth trajectory data.

At first, we perform \emph{data collection}. Specifically, given a set of environments $\mathcal{E}$, we collect the performance $P_{E}$ of \red{a given SLAM algorithm} in each environment $E \in \mathcal{E}$. In this paper, we consider indoor environments since they are largely available, characterized by stable features, and widely used to test SLAM algorithms. Note that obtaining $P_{E}$ could involve multiple runs and the availability of ground truth. In our case, we collect data in $100$ environments by using simulation (see Section \ref{S:simulations}) and we measure $P_{E}$ using a variant of the metric of~\cite{Kuemmerle2009} (see Section~\ref{S:metric}). 

After that, we extract a set of features $F_{E}$ which characterize the environment $E \in \mathcal{E}$. Note that this step, which we refer to as \emph{feature extraction} (Section \ref{S:featureextraction}), computes data directly from the ground truth (e.g., floor plans of buildings) and not from the maps obtained by the robots.

The third step is \emph{model learning}, in which a model that relates $P_{E}$ to $F_{E}$ is built. We use regression techniques to learn and represent their correlation (Section~\ref{s:modellearning}). 

Such model is used to do \emph{performance prediction}, namely to calculate the expected performance $\hat{P}_{E'}$ of \red{the target SLAM method} in an unseen environment $E'$. Our performance prediction extracts features $F_{E'}$ from $E'$ and finds $\hat{P}_{E'}$ given $F_{E'}$ exploiting the model built in the previous step (Section~\ref{s:performanceprediction}).

Finally, we can verify the accuracy of our performance predictions, comparing them against the actual measured performance (Section~\ref{sec:EXP}). 

In principle, the above sequence of steps can be repeated at regular intervals when the set $\mathcal{E}$ of visited environments grows, resulting in increasingly accurate performance predictions.

%% file: 04-Data.tex
\section{Steps of the Proposed Method}\label{sec:DATA}

In this Section, we discuss the details of the implementation of our method. 

\subsection{Performance Metric}
\label{S:metric}

As reported in Section~\ref{sec:ART}, the research community has developed several metrics to assess the performance of SLAM algorithms. In this work, we use the \emph{localization error} performance metric proposed by~\cite{Kuemmerle2009}, which measures the performance of SLAM algorithms according to their ability to accurately estimate the trajectory followed by the robot, computing the deformation energy that is required to superimpose the estimated trajectory onto the ground truth trajectory: the smaller the energy, the higher the accuracy of the reconstruction. 

This metric is chosen because of its generality and versatility: it is not restricted to any particular map representation format and can thus be used to measure the performance of any SLAM algorithm. The metric is also independent of the type of sensors mounted on the robot, making it applicable to a broad range of robot platforms.

We recall the definition of the metric as given in \cite{Kuemmerle2009}:

\begin{defn}
\label{def:error_of_single_run}
Let $x_{1:T}$ be the poses of the robot estimated by a SLAM algorithm from time step 1 to T during an exploration of environment $E$; in our case, $x_{t} \in SE(2)$, with $SE(2)$ being the special Euclidean group of order $2$. \\
Let $x_{1:T}^{*}$ be the associated ground truth poses of the robot during mapping.\\
Let $\delta_{i,j} = x_{j} \ominus x_{i}$ be the relative transformation that moves the pose $x_{i}$ onto $x_{j}$ and let $\delta_{i,j}^{*} = x_{j}^{*} \ominus x_{i}^{*}$.\\
Finally, let $\delta$ be a set of $N$ pairs of relative transformations over all the exploration, $\delta=\left\{\langle\delta_{i,j},\delta_{i,j}^{*}\rangle \right\}$. \\
The \emph{localization error} performance metric is defined as:
\begin{flalign*}
\varepsilon(\delta) &= \frac{1}{N}\sum_{i,j}(\delta_{i,j}\ominus\delta_{i,j}^{*})^{2} = \\
&=\frac{1}{N}\sum_{i,j}[trans(\delta_{i,j}\ominus\delta_{i,j}^{*})^{2} + rot(\delta_{i,j}\ominus\delta_{i,j}^{*})^{2}] = \\
&=\varepsilon_{t}(\delta)+\varepsilon_{r}(\delta),
\end{flalign*}
where $\ominus$ is the inverse of the standard motion composition operator and $trans(\cdot)$ and $rot(\cdot)$ are used to separate the translational and rotational components of the error.
\end{defn}

Authors of~\cite{Kuemmerle2009} leverage human knowledge to determine the $N$ relative transformations $\delta_{i,j}^{*}$ (and, consequently, the corresponding relative transformations $\delta_{i,j}$) composing $\delta$. 
After a run, a human operator analyzes the observations acquired by the robot (e.g., laser scans) to determine which ones refer to the same part of the environment and manually aligns them. The amount of displacement required for the alignment is stored as the ground truth of the relative transformation $\delta_{i,j}^{*}$ between the poses $x_{i}$ and $x_{j}$ from which the observations have been acquired. 
The human operator can match observations at semantically relevant places (e.g., loop closures), providing ground truth for global consistency. Clearly, this method does not efficiently scale as the numbers of observations, runs, and environments increase. 

%\red{Under the assumption of having full knowledge of ground truth trajectory we automatically compute a set of relations without any human intervention, thus automatizing the procedure for data collection.}
%However, the research community has developed over the years a number of simulation approaches that can be used to reproduce robots' behavior with very high fidelity without requiring the physical exploration of an environment. These techniques offer the ability to exactly and continuously track a robot's position throughout the entire duration of a simulation run without requiring any special equipment, and can thus be the foundation for the development of more accurate, objective and scalable approaches to SLAM evaluation.

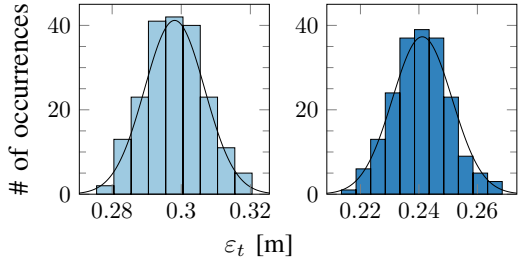
\begin{figure}
\centering
\resizebox{0.8\linewidth}{!}{
	\begin{tikzpicture}
	\begin{axis}[
		normalsize,
    	ymin=0, ymax=45,
    	xmin=0.2705,xmax=0.3255,
    	height={0.5\linewidth},
    	width={0.5\linewidth},
    	minor y tick num = 3,
    	area style,
    	ylabel = {\# of occurrences},
    	ylabel style = {yshift=-2.0ex, font=\large}
    	]
	\addplot+[hist={density},ybar interval,mark=no,black,fill=gaussianLocError_1] plot coordinates {(0.2755, 2) (0.2805, 13) (0.2855, 23) (0.2905, 41) (0.2955, 42) (0.3005, 40) (0.3055, 23) (0.3105, 11) (0.3155, 5)(0.3205,1)};
\addplot[domain={0.2705:0.3255},yscale=0.8973,samples=200] {\gauss{0.29807859}{0.008694297}};
	\end{axis}
	\end{tikzpicture}
	\begin{tikzpicture}
	\begin{axis}[
    	ymin=0, ymax=45,
    	xmin=0.2085,xmax=0.2735,
    	normalsize,
    	height={0.5\linewidth},
    	width={0.5\linewidth},
    	minor y tick num = 3,
    	area style,
    	]
	\addplot+[hist=density,ybar interval,mark=no,black,fill=gaussianLocError_2] plot coordinates { (0.2135, 1) (0.2185, 6) (0.2235, 13) (0.2285, 24) (0.2335, 37) (0.2385, 39) (0.2435, 37) (0.2485, 23) (0.2535, 9) (0.2585, 5) (0.2635,3) (0.2685,3)};
	\addplot[domain={0.2085:0.2735},yscale=0.9539,samples=200] {\gauss{0.241151215}{0.010202966}};
	\end{axis}
	\end{tikzpicture}
}\\
	\begin{tikzpicture}
    \node[below] at (0,-1.5) {$\varepsilon_{t}$ [m]};
    \end{tikzpicture}
\caption{Distribution of the translational localization error $\varepsilon_{t}$ in two environments.}
\label{fig:gaussian_approximation}
\end{figure}

We propose a new way to calculate $\delta_{i,j}^{*}$ that is independent of human intervention, enabling the possibility to collect data in a fully automatized way. In our case, the ground truth is available from simulations (Section~\ref{S:simulations}) and, in principle, $\delta$ could contain all the possible pairs of relative transformations (i.e., for all the $i$ and $j$ in $1:T$).  
%, and a significantly higher level of representativeness of SLAM performance.
%Theoretically, knowledge of the ground truth trajectory allows the straightforward computation of the localization error on the set of all possible relations. However, this approach has limited applicability due to its significant computational complexity, which is quadratic in the number of poses on the robot's trajectory. %and thus makes the metric increasingly difficult to compute as the size of the environment grows.
Our proposal is to randomly sample a set of relative transformations, the size of which is determined as a compromise between sampling accuracy and computational complexity. The procedure is based on the central limit theorem to approximate the sampling distribution with a normal distribution~\cite{billingsley1995probability}.
At first, we set the confidence level and the margin of error of the estimation. Then, we determine the number of relative transformations sampled for estimating the localization error as:
\begin{equation}\label{eq:est}
	N = \frac{z_{\alpha/2} * s^2}{d^2},
\end{equation}
where $s^2$ is the usual unbiased estimator of the population variance, $d$ is the margin of error, $\alpha$ is the complement of the desired confidence level and $z_{\alpha/2}$ is its associated z-score. 
To validate our approach, we empirically verify the distribution normality assumption on a representative set of environments. Fig.~\ref{fig:gaussian_approximation} shows the sample distribution of the translational localization error $\varepsilon_{t}$ accumulated for two of these environments; the distributions are obtained by repeatedly applying the sampling procedure to extract $200$~different samples of relative transformations imposing a $99$\,\% confidence level and a margin of error of $\pm\,0.02$~m. The shape of the distributions is approximately normal. 
%For this operation to be sound, we must assume relative transformations to be independent and identically distributed random variables. In principle, this may not be the case for every possible pair of relations, as transformations involving pairs of poses that are close to each other will inevitably be similar and not independent; however, the number of possible relations is so large that, given any two random relations, the likelihood of them being related is negligible for all practical purposes. Furthermore, as the process used for data collection is the same for all poses, it is reasonable to assume the identical distribution property to hold as well.

%% FINE PARTE DI KUMMERLE
As it stands, the above metric evaluates the performance of a SLAM algorithm over a single run. However, the performance of a SLAM algorithm in an environment may vary across multiple runs, due to noisy measurements, variations in the followed trajectory, and also a certain level of randomness that is inherent to the behavior of some SLAM algorithms.

We thus propose the following straightforward generalization of the previous definitions to model the concept of \emph{average} localization error over multiple runs of an environment.

\begin{defn}
\label{def:operational_loc_error}
Let $\mathcal{R}_{E}$ be the set of exploration runs performed on $E \in \mathcal{E}$, and $\varepsilon_{t}(\delta_{R_{E}})$ and $\varepsilon_{r}(\delta_{R_{E}})$ the translational and rotational errors measured over run $R_{E} \in \mathcal{R}_{E}$.\\
The \emph{mean and standard deviation of the translational localization error} of $E$ are defined as:
\begin{flalign*}
\mean{\varepsilon_{t}(E)} &= \frac{\sum\limits_{R \in \mathcal{R}_{E}} \varepsilon_{t}(\delta_R)}{\left\vert{\mathcal{R}_{E}}\right\vert}\\\\
s(\varepsilon_{t}(E)) &= \sqrt {\frac{\sum\limits_{R \in \mathcal{R}_{E}} [\varepsilon_{t}(\delta_R)-\mean{\varepsilon_{t}(E)}]^2}{\left\vert{\mathcal{R}_{E}}\right\vert}}.
\end{flalign*}
The \emph{mean and standard deviation of the rotational localization error} of $E$, $\mean{\varepsilon_{r}(E)}$ and  $s(\varepsilon_{r}(E))$, are defined accordingly.
%\begin{flalign*}
%\mean{\varepsilon_{r}(E)} &= \frac{\sum\limits_{R \in \mathcal{R}_{E}} \varepsilon_{r}(\delta_R)}{\left\vert{\mathcal{R}_{E}}\right\vert}\\\\
%s(\varepsilon_{r}(E)) &= \sqrt {\frac{\sum\limits_{R \in \mathcal{R}_{E}} [\varepsilon_{r}(\delta_R)-\mean{\varepsilon_{r}(E)}]^2}{\left\vert{\mathcal{R}_{E}}\right\vert}}
%\end{flalign*}
\end{defn}

In summary, given an environment $E$, the performance $P_{E}$ is the vector $P_{E} = \langle \mean{\varepsilon_{t}(E)}, s(\varepsilon_{t}(E)), \mean{\varepsilon_{r}(E)}, s(\varepsilon_{r}(E)) \rangle$.
%The reasons behind this choice are twofold. First, as it only requires ground truth and estimated trajectory data to be computed, it does not rely on any particular map format; this makes it a very versatile metric, as it can be applied to any SLAM algorithm independently on the chosen map representation. Second, it does not require any specific  

\subsection{Sample Size Estimation}
\label{sample_size_estimation}

%In order to obtain reliable values for $\mean{\varepsilon_{t}(E)}$, $s(\varepsilon_{t}(E))$, $\mean{\varepsilon_{r}(E)}$, $s(\varepsilon_{r}(E))$, we need a criterion to identify how many runs $|\mathcal{R}_{E}|$ should be performed for each environment $E$. %To do so, we adopt a technique for the estimation of the sample size that is conceptually similar to the one we used in Section \ref{ssec:single_exploration_metric} to determine the number of ground truth relations for an individual run.

In principle, we cannot assume that \red{SLAM performances obtained by the same algorithm in different runs performed} in the same environment to be completely independent of each other. However, Chebyshev's weak law of large numbers guarantees the convergence of the sample mean to the true mean under the assumption that the covariances tend to be zero on average~\cite{taylorfirst}. In order to obtain reliable values for $\mean{\varepsilon_{t}(E)}$, $s(\varepsilon_{t}(E))$, $\mean{\varepsilon_{r}(E)}$, $s(\varepsilon_{r}(E))$, we therefore need a criterion to identify how many runs $|\mathcal{R}_{E}|$ should be performed for each environment $E$. For this purpose, we assume the distribution of the sample mean to be approximately normal and we use the same formulation of Equation \eqref{eq:est} to obtain $|\mathcal{R}_{E}|$.

%In principle, we cannot assume different runs of GMapping in the same environment to be completely independent from each other. 
%; this is because, despite them being conducted separately and with no mutual interference, they might be performed according to the same exploration criteria and thus have similar trajectories. 
%We therefore have to assume their covariances to be sufficiently small to still apply the Chebyshev's weak law of large numbers for potentially correlated sequences. 
%This also means that the distribution of the sample mean may not be normal, as the central limit theorem is not guaranteed to hold. 
%For the purpose of determining a viable number of sample exploration runs, 
%Then, we assume the distribution of the sample mean to be approximately normal and we use the same formulation of Equation \eqref{eq:est} to obtain $|\mathcal{R}_{E}|$. % QUESTA PARTE COMMENTATA E' UNA RIPETIZIONE, FORSE POSSIAMO TOGLIERLA? 
%, with $s^2$ is the usual unbiased estimator of the population variance, $d$ is the margin of error, $\alpha$ is the complement of the desired confidence level and $z_{\alpha/2}$ is its associated z-score.

The sample size estimation process is bootstrapped with an estimate of the variance of the localization error, which we obtain with a pilot search on a small sample of $10$~runs. We use this value to compute an initial estimate of the number of required runs. We then perform that number of runs and compute a new estimate of the variance and its associated sample size, iteratively repeating the process until the newly estimated sample size is not larger than the number of already performed runs. In our case, we end up with an average of $|\mathcal{R}_{E}|=36$ runs for each environment $E$.

%Note that the sample size $n_{t}$ required for an accurate estimation of the translational localization error may differ from the sample size $n_{r}$ required to accurately estimate the rotational localization error; the selected sample size is the maximum of $n_{t}$ and $n_{r}$.

\subsection{Data Collection}
\label{S:simulations}

As the set of environments $\mathcal{E}$ over which we collect data on the performance of \red{SLAM algorithms}, we consider a set of $100$ indoor environments covering a wide range of building types (schools, offices, university campuses, and others), sizes, and shapes, to avoid overfitting. 
%More precisely, we collect data on performance of Gmapping in $105$ real world buildings.
% stored in the \texttt{png} image format and have a resolution of $0.05$ meters per grid cel.
%VALERIO: DIRE QUANTI NE PRENDIAMO E DA QUALI
Some environments are selected from those used in \cite{7487234}, which include floor plans from the Radish repository. Such environments range from $\SI[per-mode = symbol]{100}{\meter\squared}$ to $\SI[per-mode = symbol]{1000}{\meter\squared}$, and are equally divided between offices and research laboratories. We select a subset of $11$ of the $20$ environments of~\cite{7487234} for which the ground truth floor plans are available and we limit our analysis to their empty (without furniture) version.
We also use a set of $25$ floor plans that represent buildings of the MIT university campus. These floor plans are part of the dataset used in~\cite{whiting2007generating} and comprise research laboratories, offices, and teaching areas, ranging from about $\SI[per-mode = symbol]{1000}{\meter\squared}$ to over $\SI[per-mode = symbol]{30000}{\meter\squared}$.
Finally, we use our own dataset of $64$ floorplans representing real-world buildings~\cite{robocup2013}, with $26$ offices and $38$ schools, ranging from about $\SI[per-mode = symbol]{100}{\meter\squared}$ to over $\SI[per-mode = symbol]{10000}{\meter\squared}$.
\red{In all of the performed runs, data are collected when the robot covered the entirety of the environment, so when the map represents the whole environment. Incomplete runs as in the case where one of the exploration nodes experienced a failure or the robot got stuck are automatically discarded and a new run is executed instead.}
\red{While performances of Gmapping have been collected on all of $100$ environments, we collected data regarding KartoSLAM on a subset of $80$ of those, due to different requirements for KartoSLAM to work properly in such environments. }  

We use simulations on environments $E \in \mathcal{E}$ to collect data on performance $P_{E}$. Simulations allow us to automatically execute multiple runs for each environment and on a large set of environments at a significantly lower cost than what would be possible with a real robot. In addition, simulations naturally provide the ground truth trajectory data that is required for the evaluation of the localization error. We validate our simulations in Section~\ref{sec:EXP}.

Simulations are performed with Stage, using the Navigation packages\footnote{\url{http://wiki.ros.org/navigation}} robot movement and the ROS GMapping\footnote{\url{http://wiki.ros.org/gmapping}} and KartoSLAM\footnote{\url{http://wiki.ros.org/slam_karto}} for SLAM. We consider a single virtual robot equipped with a two-dimensional laser range scanner having a field of view of $\SI[per-mode = symbol]{270}{\degree}$, an angular resolution of $\SI[per-mode = symbol]{0.5}{\degree}$, %a $[0, 60]$ $\SI[per-mode = symbol]{}{\meter}$ range, 
and a maximum range of $\SI[per-mode = symbol]{30}{\meter}$. We assume the virtual robot has a translational odometry error of $\SI[per-mode = symbol]{0.01}{\meter\per\meter}$ and a rotational odometry error up to $\SI[per-mode = symbol]{2}{\degree\per\radian}$, which is considered a reasonable approximation of the odometry accuracy of a real wheeled robot.

Given an environment $E$, and a starting pose for the robot (close to the center of the environment, the same for all runs), a run $R_{E}$ evolves by exploring $E$ using the frontier-based exploration paradigm defined in~\cite{613851}, where each frontier represents a region on the boundary between open space and unexplored space.
% an area of the space is considered to be \emph{explored} once it becomes part of the map produced by the SLAM algorithm. %To make the exploration process more robust, points that are located at very short distance from each another are clustered in a single frontier.
%
%In order to select which frontier should be explored after a goal has been reached, frontiers are ordered according to their euclidean distance from the current position of the robot; the goal is then chosen on the frontier that is nearest to the robot, with eventual ties broken randomly.
%
%As said in Section \ref{ssec:single_exploration_metric}, our approach to consider an
%Q: DOBBIAMO DIRE IL VALORE ESATTO DI SIMILARITY?NO
A run $R_{E}$ is complete when two snapshots of the map produced by the SLAM algorithm (taken at regular time intervals) are similar enough, according to the mean square error metric to evaluate the difference between images. Once $R_{E}$ is complete, its localization error ($\varepsilon_{t}(\delta_{R_{E}})$ and $\varepsilon_{r}(\delta_{R_{E}})$) is computed. %The process is repeated until the number of runs $|\mathcal{R}_{E}|$ performed on an environment $E$ becomes less than or equal to the estimated sample size for that environment, as we discussed in the previous section

\red{
At the end of each run $R_{E}$, we have thus the set $D(R_{E})$ of data (laser range scans and odometry readings) fed to SLAM method, and its output $M(R_{E})$ (the grid map) and $x_{1:T_{R_{E}}}$ (the estimated poses of the robot from time step $1$ to $T_{R_{E}}$, the time step at which the exploration run $R_{E}$ ended). The process is automatically iterated until a number of runs $|\mathcal{R}_{E}|$ (where $\mathcal{R}_{E}$ is the set of runs performed in $E$) are performed for each environment $E$, as we discuss in Section~\ref{sample_size_estimation}. Eventually, for each environment $E  \in \mathcal{E}$, we have the set $\mathcal{D}_{E} = \{ D(R_{E}) \text{ for all } R_{E} \in \mathcal{R}_{E} \}$ of test data and the corresponding results produced by the SLAM method, namely the set of grid maps $\{ M(R_{E}) \text{ for all } R_{E} \in \mathcal{R}_{E} \}$ and the set of estimated poses $\{ x_{1:T_{R_{E}}} \text{ for all } R_{E} \in \mathcal{R}_{E} \}$. (We note that the test data $\mathcal{D}_{E}$ could be used to evaluate other SLAM algorithms without the need to re-run all simulations, but only those needed to reach $|\mathcal{R}_{E}|$ for that SLAM method.)
We point out that the test data $\mathcal{D}_{E}$ are relative to the particular configuration of the virtual robot (and of its sensors) that we have considered. For example, changing the field of view or the range of the laser range scanner leads to generating a set of data that could be different. However, the use of Stage allowed us to obtain easily and with a reduced computational effort and with limited supervision the required data thus preserving the realism of the simulated setting \cite{OURIROS}; consequently generating a new set of test data $\mathcal{D}_{E}$ for an environment $E$ with a new robot configuration is relatively cheap. For example, in the largest environment we considered, an exploration run requires, on average, $43.8$ minutes. Similarly, the data $\{ M(R_{E}) \}$ and $\{ x_{1:T_{R_{E}}} \}$, representing the results of the SLAM method, depend on the configuration of the algorithm (and, of course, on test data $\mathcal{D}_{E}$). If needed, the specific configuration of a different robotic platform or of a different sensor can be easily modeled and a new dataset with such configuration can be obtained semi-automatically.
}

\begin{figure}[t!]
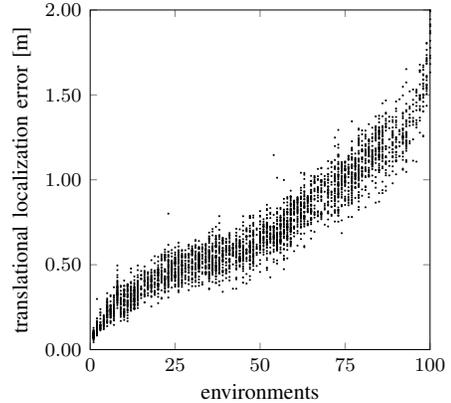

  \centering
  \resizebox{0.7\linewidth}{!}{

	}
\caption{\red{The translational localization errors for all the runs in our environments using the GMapping as SLAM method where we automatically compute the number runs needed (for each environment) to estimate the environmment localization error and the number of relatios needed (for each run) for computing the run localization error.} }
\label{F:allruns-allenvironments}
\end{figure}

Overall, we perform more than $3500$ simulated runs in Stage for each SLAM method; \red{Fig. \ref{F:allruns-allenvironments} shows the localization errors of $\mathcal{E}$  obtained with GMapping for all of the environments considered (the rotational localization errors are similar, as well as data obtained with KartoSLAM).}

%DARE ALLA FINE IL NUMERO TOTALE (at the end we collected 3600 dataset o qualcosa cosi').

%% file: 05-Method.tex
%\section{Method}\label{sec:MAIN}
%{\color{red}Sezione principale dove viene spiegato il metododo. 
%ambiente -> floor plan -> feature da una parte. Dall'atra invee abbiamo ambiente->sensor->slam -> map -> metric. Il flow chart del metdodo deve essere ben chiaro, la metrica dell'errore a questo punto è stata già introdotta in Sezione \ref{sec:DATA}.

%Descrizione qui delle feature, di come vengono estratte e di come è fatto il modello di regressione.}

\subsection{Feature Extraction}
\label{S:featureextraction}

\red{Let $\mathbb{E}$ be the set of possible environments in which a SLAM algorithm can be run. A \emph{feature} is a function $f\in\mathbb{F}: \mathbb{E} \mapsto \mathbb{R}$ that, given any environment $E \in \mathbb{E}$, returns a real number that denotes some property of $E$. In this work, we identify a set of features that, for each environment $E$, describes in an explicit and analytical method the environment complexity and which is highly correlated to the difficulty encountered by a robot to map and to localize itself within that environment. The main idea behind this is that the more the structure of a building is complex, the bigger the building is, the more difficult the robot's operations are. However, we empirically discovered that features that were representative of the building complexity were not highly correlated with the final map quality and localization performances, as not only the building complexity itself has to be taken into account, but also the complexity of the actions that the robot needs to perform to operate in such a context. }

Following these insights, we extract a set of features $F_{E}$ that characterize an environment $E$ using the Voronoi graph built on $E$ \cite{Schwertfeger2016}. Intuitively, the edges of the Voronoi graph represent the clearest paths (i.e., the paths furthest from obstacles) that a robot could follow in the environment. We exploit the Voronoi graph to obtain an approximation of the path a robot follows to build the complete map of an initially unknown environment, considering the limitations of the robot sensors. 

\begin{figure}[t!]
\centering
\begin{subfloat}
  \centering
  \includegraphics[width=0.43\linewidth]{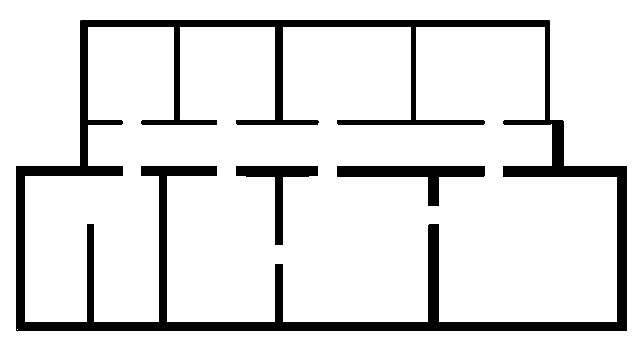}
\end{subfloat}%
\begin{subfloat}
  \centering
  \includegraphics[width=0.43\linewidth]{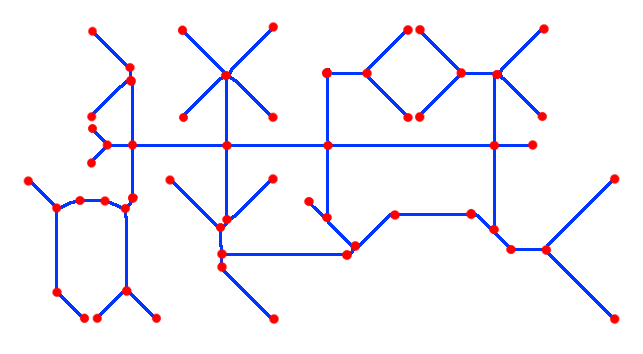}
\end{subfloat}%
\\
\caption{A floor plan (left) and the corresponding Voronoi graph (right; cells are red and blue pixels).}
\label{fig:voronoi_steps}
\end{figure}

In practice, given the floor plan of $E$, represented as a grid map (image), we extract the Voronoi graph from the dual Delaunay triangulation, as specified in~\cite{doi:10.1093/comjnl/24.2.162}. 
%We also employ the longest contour identified with the algorithm of~\cite{SUZUKI198532} to divide the inside and the outside of each environment $E$ .
\red{In order to distinguish the inside of each building from its outside, we identify the boundaries of the floor plan as the longest contour identified by the algorithm of \cite{SUZUKI198532}; we, therefore, check each pixel of the map image and we retain only those pixels that fall inside the contour. At first, we apply an image dilation operator to the map after Voronoi tesselation using a $5\times5$ pixels kernel; afterward, we reduce the resulting image to a $1$ pixel wide representation using the skeletonization operator defined in \cite{Zhang:1984:FPA:357994.358023}. These two steps are necessary to obtain a representation of the map after Voronoi tesselation that is as clean and uniform as possible and to remove any imperfection and irregularity that may have been introduced during tesselation. We consider each pixel as a node of the Voronoi graph, adding an edge of unitary weight between two nodes if their corresponding pixels are \emph{close} to each other, i.e., if one sits within the $3\times3$ pixel area centered around the other. To reduce the size of the graph and obtain something more manageable, we then perform a graph sparsification that removes all \emph{pass-through} nodes, i.e, nodes that have exactly two neighbors and whose corresponding pixels belong to a straight line; the neighbors are then connected via a new edge whose weight is the sum of the weights of the original edges it's replacing.}
\blue{The Voronoi graph obtained in this way is represented as a set of cells (pixels) and superimposed to the grid map see Fig.~\ref{fig:voronoi_steps}.}

Given the Voronoi graph of an environment $E$, we calculate two quantities, the \emph{Voronoi traversal distance} ($\textit{VTD}_{E}$) and \emph{Voronoi traversal rotation} ($\textit{VTR}_{E}$), that are the overall distance and the overall amount of rotation, respectively, that a robot incurs to cover the Voronoi graph. Specifically, imagine a virtual robot that moves on the cells composing the Voronoi graph and that perceives the environment with a sensor with a given range and field of view (we use the same configuration as in Section \ref{S:simulations}). The robot starts from the cell of the Voronoi graph closest to its actual starting position in $E$. Then, the robot marks as viewed the cells of the Voronoi graph that are visible from its current cell, it identifies the nearest unexplored cell on the Voronoi graph, and it moves toward it along the shortest path (perceiving also while traveling). The exploration ends when all the cells of the Voronoi graph have been marked as viewed. \red{Our proposal is to use the Voronoi graph of the environment to obtain an approximation of the path the robot would follow to perform a complete exploration of that environment by simulating a mapping and exploration process on the Voronoi graph and on the floor plan of $E$.  During the mapping phase, the algorithm identifies which nodes of the Voronoi graph correspond to pixels of the map that are \emph{visible} from the robot's pose and uses this information to update its internal representation of which portions of the environment have been mapped so far. Intuitively, visible pixels represent areas of the environment that the robot would be able to perceive with a laser scan from its current pose, and that would consequently become part of the map maintained by the SLAM algorithm.}
\red{ To discover such pixels, the algorithm performs a ray casting operation on the superimposition of the Voronoi graph and the floor plan's image from the current robot's position. A pixel is considered to be visible if it is within the laser's range and field of view and if there are no obstacles, black pixels on the floor plan's image, that belong to the straight line connecting it to the current robot location. In the exploration phase, the algorithm identifies the node of the Voronoi graph that corresponds to the next frontier to explore and moves the virtual robot towards it, accounting for the areas of the environment that get explored while moving. It then uses Dijkstra's algorithm to compute the shortest path on the Voronoi graph that connects the node of the robot's position to the node that corresponds to the selected map pixel. For each node along the path, the algorithm uses image map to update the virtual robot's position and orientation according to the node's relative displacement from its immediate predecessor on the path; it then adds the amount of performed translation and rotation to the overall Voronoi traversal distance and Voronoi traversal rotation respectively. Finally, the algorithm performs a mapping phase to identify which pixels would be visible from that pose and updates the \emph{seen} status of their corresponding nodes accordingly, in order to keep track of which areas of the environment have been explored.} 
\red{ The Voronoi traversal distance is estimated by computing the Euclidean distance between the pixels of the image map graph corresponding to the two nodes, which is a good approximation of the real distance the robot would have to travel to move between them.}
\red{ On the other hand, the Voronoi traversal rotation is estimated by computing the absolute value of the minimum angular difference between the current orientation of the robot and the gradient of the straight line connecting the two nodes, but only if the two nodes are at a minimum distance from each other. This adjustment is made necessary by the consideration that the relative orientation difference between two nearby pixels does not correctly reflect the actual amount of rotation the robot performs along the trajectory and is instead strongly influenced by imperfections in the alignment of the pixels; therefore, evaluating the relative orientation of two nodes only if their distance is sufficiently large represents a better approximation of the actual amount of rotation the robot would perform between them, as it only accounts for rotations in the correspondence of curves in the simulated trajectory.}
%\textcolor{blue}{Giustificare la scelta di feature di voronoi con il fatto che, di fatto, proviamo a stimare un path medio rispetto ai vari path possibili dell'ambiente, che tocchi tutti i nodi}
At the end of this simulated exploration, the total amount of translation and rotation are the values of $\text{VTD}_{E}$ and $\text{VTR}_{E}$, respectively, that approximate the amount of distance and rotation that a robot has to perform in order to map $E$. Feature extraction for an environment $E$ takes less than $4$ minutes on a commercial laptop.

In summary, the features of $E$ we consider in the following are $F_{E} = \langle \textit{VTD}_{E}, \textit{VTR}_{E}\rangle$.
\red{Full details of the feature extraction mechanism and our code are publicly available}\footnote{\url{https://github.com/AIRLab-POLIMI/predictivebenchmarking}.}

\red{Besides these features we have evaluated the use of several other features that represent the structural complexity and the geometrical complexity of an environment, to assess the extent to which these features could be used for estimating the mapping and localization performances. Such features were computed starting from the floorplan of an environment using the method of \cite{OURIAS}. More precisely, we considered the total area of the building, the perimeter of the building, the ratio between the longest and the shortest external wall, the number of rooms, and the total sum of the perimeter of each room. We also extracted a topological graph representing the building, where each room is considered as a node and an edge is placed when two rooms are connected. This topological graph $G$ is used for computing features related to the complexity of the topological graph. More precisely we computed the order, size, density, and radius of $G$. We also computed the number of \emph{bifurcation points} and \emph{terminal points} of $G$, its \emph{density}, and several graph centrality measures \cite{newman2003mixing} (Betweenness Centrality, Katz Centrality, Closeness Centrality, Eigenvector Centrality).}
%\blue{TROVARE CITAZIONE PER CENTRALITY E DIRE CHE NON ABBIAMO USATO I DATI DEL TRAINING PER SCELTA}

\subsection{Model Learning}\label{s:modellearning}

In this step, we build a model of the relationships between performance $P_{E}$ and features $F_{E}$ for all the environments $E \in \mathcal{E}$. 
To assess the quality of a possible model, we compute the overall root mean square error (RMSE) and the average coefficient of determination ($R^2$) on the training set $\mathcal{E}$ using k-fold cross validation (with a number of folds equal to $5$). We also consider the normalized (scale-independent) version of the RMSE: $
\text{NRMSE} = \frac{\text{RMSE}}{y_{\text{max}}-y_{\text{min}}}$, where $y_{\text{max}}$ and $y_{\text{min}}$ are the maximum and the minimum observed values of the target variable $y$, respectively.

A distinct model is built for each one of the four components of $P_{E} = \langle \mean{\varepsilon_{t}(E)}, s(\varepsilon_{t}(E)), \mean{\varepsilon_{r}(E)}, s(\varepsilon_{r}(E)) \rangle$. Surprisingly, we find that the two features of $F_{E}$ linearly correlate to the components of $P_{E}$ and the corresponding models can be built performing linear regression.

\begin{figure*}[h]
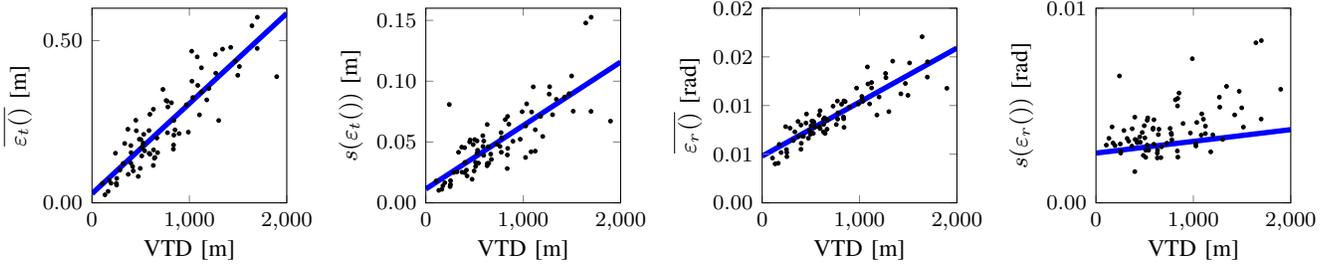

\centering
\begin{subfloat}{
	%
}
\end{subfloat}%
\caption{The regression lines representing the models that correlates the Voronoi Traversal Distance (VTD) with the four components of the localization error (black dots are values relative to individual environments in $\mathcal{E}$) \red{for maps obtained using the KartoSLAM algorithm.}}
\label{fig:voronoi_traversal_distance_fitness_KARTO}
\end{figure*}

Table \ref{tab:best_single_feature} and Fig.~\ref{fig:voronoi_traversal_distance_fitness} shows the best model, in terms of highest $R^2$ value \red{for maps obtained with GMapping as SLAM method, } 
%Each model is also accompanied by its $5$-fold cross validated RMSE together with its normalized variant adjusted for scale-independence. 
\red{while Table \ref{tab:best_single_feature_KARTO} and Fig.~\ref{fig:voronoi_traversal_distance_fitness_KARTO} show the results obtained while predicting the quality of maps obtained KartoSLAM using a single feature. }
%For additional context, Table \ref{tab:second_best_single_feature} shows the second best performing single-feature linear model.
\red{For each of the four components of the localization error we are interested in predicting, and for all SLAM algorithm considered, $\text{VTD}_{E}$ correlates to all the components of $P_{E}$ better than $\text{VTR}_{E}$.}

\begin{table}[t!]
\centering
\caption{Performance of the best single-feature linear models for the components of the localization error based on features $F_{E}$ \red{when considering the GMapping SLAM method}. NRMSE is expressed in percentage with respect to the range of the predicted variable, while RMSE is expressed in m for the translations and in rad for the rotations.}
\label{tab:best_single_feature}
\begin{tabular}{l|l|c|c|c|}
\cline{2-5}
 & \multicolumn{1}{c|}{feature} & \multicolumn{1}{c|}{$R^{2}$} & \multicolumn{1}{c|}{RMSE} & \multicolumn{1}{c|}{NRMSE} \\ \hline 
\multicolumn{1}{|l|}{$\mean{\varepsilon_{t}()}$} & VTD & \multicolumn{1}{c|}{0.830} & \multicolumn{1}{c|}{0.145} & \multicolumn{1}{c|}{7.92\%} \\ \hline
\multicolumn{1}{|l|}{$s(\varepsilon_{t}())$} & VTD & 0.621 & 0.018 & 9.87\% \\ \hline
\multicolumn{1}{|l|}{$\mean{\varepsilon_{r}()}$} & VTD & 0.723 & 0.004 & 11.15\% \\ \hline
\multicolumn{1}{|l|}{$s(\varepsilon_{r}())$} & VTD & 0.091 & 0.002 & 14.85\% \\ \hline
\end{tabular}%
\end{table}

\begin{table}[t!]
\centering
\caption{Performance of the best single-feature linear models for the components of the localization error based on features $F_{E}$ \red{when considering the KartoSLAM method}. See Table \ref{tab:best_single_feature} for the notation.}
\label{tab:best_single_feature_KARTO}
\begin{tabular}{l|l|c|c|c|}
\cline{2-5}
 & \multicolumn{1}{c|}{feature} & \multicolumn{1}{c|}{$R^{2}$} & \multicolumn{1}{c|}{RMSE} & \multicolumn{1}{c|}{NRMSE} \\ \hline 
\multicolumn{1}{|l|}{$\mean{\varepsilon_{t}()}$} & VTD & \multicolumn{1}{c|}{0.763} & \multicolumn{1}{c|}{0.063} & \multicolumn{1}{c|}{11.54\%} \\ \hline
\multicolumn{1}{|l|}{$s(\varepsilon_{t}())$} & VTD & 0.544 & 0.093 & 12.15\% \\ \hline
\multicolumn{1}{|l|}{$\mean{\varepsilon_{r}()}$} & VTD & 0.741 & 0.001 & 9.68\% \\ \hline
\multicolumn{1}{|l|}{$s(\varepsilon_{r}())$} & VTD & 0.622 & 0.0006 & 16.47\% \\ \hline
\end{tabular}%
\end{table}
The low $R^2$ value for the best model predicting the standard deviation of the rotational localization error $s(\varepsilon_{r}())$ suggests that none of the features we have considered fits the data particularly well for predicting $s(\varepsilon_{r}())$. 
%as their performance is not significantly better than that of the mean predictor. 
However, the analysis of the RMSE and of the NRMSE reveals that the average prediction accuracy is actually quite good, with an RMSE of less than $\SI{0.002}{\radian}$. 
One of the reasons for this apparent contradiction is that the standard deviation of the rotational localization error shows little variability across all the environments of the training set $\mathcal{E}$, as it can be seen from the scale of \red{ both Fig.~\ref{fig:voronoi_traversal_distance_fitness} and Fig.~\ref{fig:voronoi_traversal_distance_fitness_KARTO}} right, for which the effects of random noise become visible.

%\begin{figure}[H]
%\centering
%\begin{subfigure}
%  \centering
%  \includegraphics[width=.43\linewidth]{pictures/cap7/sim_models/realistic/individual/transError_mean_mean/voronoi_traversal_distance.png}
%  \caption{Translational error mean}
%  \label{fig:voronoi_traversal_distance_fitness_sub1}
%\end{subfigure}%
%\begin{subfigure}
%  \centering
%  \includegraphics[width=.43\linewidth]{pictures/cap7/sim_models/realistic/individual/transError_mean_std/voronoi_traversal_distance.png}
%  \caption{Translational error std. dev.}
%  \label{fig:voronoi_traversal_distance_fitness_sub2}
%\end{subfigure}%
%\\
%\begin{subfigure}
%  \centering
%  \includegraphics[width=.43\linewidth]{pictures/cap7/sim_models/realistic/individual/rotError_mean_mean/voronoi_traversal_distance.png}
%  \caption{Rotational error mean}
%  \label{fig:voronoi_traversal_distance_fitness_sub3}
%\end{subfigure}%
%\begin{subfigure}
%  \centering
%  \includegraphics[width=.43\linewidth]{pictures/cap7/sim_models/realistic/individual/rotError_mean_std/voronoi_traversal_distance.png}
%  \caption{Rotational error std. dev.}
%  \label{fig:voronoi_traversal_distance_fitness_sub4}
%\end{subfigure}
%\caption{The regression line of the VTD model for the four components of the localization error on the training environments. The x axis represents the VTD of the environments, the y axis represents the value of the localization error, and the black dots show the true SLAM performance on the training environments.}
%\label{fig:voronoi_traversal_distance_fitness}
%\end{figure}

\begin{figure*}
\centering
\begin{subfloat}{
}
\end{subfloat}%
\caption{The regression lines representing the models that correlates the Voronoi Traversal Distance (VTD) with the four components of the localization error (black dots are values relative to individual environments in $\mathcal{E}$) \red{for maps obtained using the GMapping SLAM algorithm}.}
\label{fig:voronoi_traversal_distance_fitness}
\end{figure*}

Other interesting insights come from the comparison between the localization error predicted by the VTD-based model and that predicted using the actual values of the travelled distance and rotation measured along the actual trajectories followed by the robot, shown in Table \ref{tab:best_true_feature}. (Note that the model based on actual trajectory data cannot be used for prediction.)
% and in Figures \ref{fig:true_length_predictor_fitness} and \ref{fig:true_rot_predictor_fitness}. 

\begin{table}[H]
\centering
\caption{Performance of the best single-feature linear models for components of the localization error based on actual trajectory data.}
\label{tab:best_true_feature}
\begin{tabular}{l|l|c|c|c|}
\cline{2-5}
 & \multicolumn{1}{c|}{feature} & \multicolumn{1}{c|}{$R^{2}$} & \multicolumn{1}{c|}{RMSE} & \multicolumn{1}{c|}{NRMSE} \\ \hline
\multicolumn{1}{|l|}{$\mean{\varepsilon_{t}()}$} & trajectory length & \multicolumn{1}{c|}{0.905} & \multicolumn{1}{c|}{0.116} & \multicolumn{1}{c|}{6.36\%} \\ \hline
\multicolumn{1}{|l|}{$s(\varepsilon_{t}())$} &  trajectory length & 0.469 & 0.023 & 12.38\% \\ \hline
\multicolumn{1}{|l|}{$\mean{\varepsilon_{r}()}$} &  trajectory rotation & 0.812 & 0.003 & 8.55\% \\ \hline
\multicolumn{1}{|l|}{$s(\varepsilon_{r}())$} &  trajectory rotation & 0.081 & 0.002 & 15.28\% \\ \hline
\end{tabular}%
\end{table}

%\textcolor{red}{Il commento 3 del revisore 3 dice che output del training process non è usato per il modello. E' vero, ma è vero anche che siccome il modello lineare è molto accurato, la R2 ottenuta dal nostro modello, seppur semplice, è 
%praticamente pari a quella di usare delle misure ex-post come la lunghezza del path e uesto giustifica il nostro modello. In pratica, è vero che non usiamo i training data ma usiamo solo le features estratte direttamente dal GT, ma perché le nostre features estratte direttamente dal GT sono significative praticamente come delle informazioni che sono ottenibili solo andando ad eseguire dei run nell'ambiente, come si vede in Table II.}
The results show that the actual trajectory data are strongly correlated with the measured localization error, with the exception of the standard deviation of the rotational component, $s(\varepsilon_{r}(E))$, that is substantially uncorrelated, as before. These data confirm that the amount of localization error committed by GMapping in an environment is mostly dependent on just travelled distance and rotation. Although this claim sounds intuitive, we  provide empirical data to quantitatively support it. We believe one of the reasons behind this result is that longer trajectories in indoor environments typically involve the traversal of long corridors or large spaces, which are relatively featureless and may result in reducing the capability of GMapping to exploit the characteristics of the environment to perform localization. A similar consideration may hold for the increase of the overall amount of rotation, which is most likely associated with the robot entering and exiting a large number of rooms, eventually decreasing the localization capability of GMapping.

Finally, if we compare the model based on the actual trajectory data with the model based on VTD, we can see that the two $R^2$ coefficients are very close to each other; this implicitly suggests that the VTD feature is indeed a good predictor of the actual distance travelled by the robot.

\subsection{Other Models}\label{s:modelother}
\red{ While we report here only the linear regression models based on VTD and VTR, we have considered several other features for $F_{E}$, as mentioned in Section \ref{S:featureextraction}, as the building perimeter, its area, and the number of nodes and edges of the Voronoi graph. Moreover, we have investigated if multi-variate linear models and non-linear models. The results obtained are and the techniques used are briefly described here.}

\red{ The first approach we investigate is multiple linear regression, i.e., a linear regression technique that uses more than one explanatory variable~\cite{Bishop:2006:PRM:1162264}. We use \emph{F-regression}~\cite{kutner2005applied} univariate linear regression test for selecting the $K$ features with the highest F-scores, with $K$ ranging from one to the overall number of features, and keeping the model with the best performances.}

\red{ The second approach we use is \emph{ElasticNet}, a regularized multiple linear regression method that directly performs feature selection as part of the training process that linearly combines the L1 and L2 penalties of the \emph{lasso} and \emph{ridge} regression methods to prevent overfitting~\cite{Bishop:2006:PRM:1162264,Zou05regularizationand}.} 
\red{ Finally, we use Gaussian Process (GP) regression method \cite{rasmussen2003gaussian, Bishop:2006:PRM:1162264},  
a non-linear kernel-based that defines a Gaussian posterior distribution over target functions, whose mean is used for prediction.}
\red{However, we discovered a limited advantage in using other models than linear regression based on VTD and VTR, at the expense of having a more complex and less interpretable model. More precisely, the use of multiple linear regression and elastic net provided only a minor performance improvement, within statistical significance. Details and full results are omitted for brevity but can be found, among the code used at the aforementioned repository. Tables \ref{tab:best_single_feature_GP} and \ref{tab:best_single_feature_GP_KARTO} shows the best model obtained with Gaussian Process regression, in terms of highest $R^2$ value for maps obtained with GMapping  and KartoSLAM respectively. While the performances of the model using GP regression are slightly better than linear regression, we believe that the benefit of having a linear and interpretable model constitutes a bigger advantage for real-world applications if compared to the small improvement in results obtained by GP. For this reason, in the rest of this work, we consider linear regression using VTD and VTR as the reference model. }

%\blue{DECIDERE SE METTERE O MENO DISCUSSIONE SU ELASTICNET E FEATURE SELECTION - FIG7.7 TESI VALERIO}

\begin{table}[t!]
\centering
\caption{DATI Gaussian Process Gmapping.}
\label{tab:best_single_feature_GP}
\begin{tabular}{l|l|c|c|c|}
\cline{2-5}
 & \multicolumn{1}{c|}{feature} & \multicolumn{1}{c|}{$R^{2}$} & \multicolumn{1}{c|}{RMSE} & \multicolumn{1}{c|}{NRMSE} \\ \hline 
\multicolumn{1}{|l|}{$\mean{\varepsilon_{t}()}$} & VTD & \multicolumn{1}{c|}{0.834} & \multicolumn{1}{c|}{0.141} & \multicolumn{1}{c|}{7.56\%} \\ \hline
\multicolumn{1}{|l|}{$s(\varepsilon_{t}())$} & VTD & 0.723 & 0.012 & 9.41\% \\ \hline
\multicolumn{1}{|l|}{$\mean{\varepsilon_{r}()}$} & VTD & 0.735 & 0.004 & 9.5\% \\ \hline
\multicolumn{1}{|l|}{$s(\varepsilon_{r}())$} & VTD & 0.0159 & 0.002 & 17.59\% \\ \hline
\end{tabular}%
\end{table}

\begin{table}[t!]
\centering
\caption{DATI Gaussian Process KARTO.}
\label{tab:best_single_feature_GP_KARTO}
\begin{tabular}{l|l|c|c|c|}
\cline{2-5}
 & \multicolumn{1}{c|}{feature} & \multicolumn{1}{c|}{$R^{2}$} & \multicolumn{1}{c|}{RMSE} & \multicolumn{1}{c|}{NRMSE} \\ \hline 
\multicolumn{1}{|l|}{$\mean{\varepsilon_{t}()}$} & VTD & \multicolumn{1}{c|}{0.821} & \multicolumn{1}{c|}{0.055} & \multicolumn{1}{c|}{9.97\%} \\ \hline
\multicolumn{1}{|l|}{$s(\varepsilon_{t}())$} & VTD & 0.480 & 0.018 & 13.06\% \\ \hline
\multicolumn{1}{|l|}{$\mean{\varepsilon_{r}()}$} & VTD & 0.801 & 0.001 & 8.61\% \\ \hline
\multicolumn{1}{|l|}{$s(\varepsilon_{r}())$} & VTD & 0.196 & 0.0006 & 16.77\% \\ \hline
\end{tabular}%
\end{table}

\subsection{Performance Prediction}\label{s:performanceprediction}

The last step of our method predicts the performance $\hat{P}_{E'}$ \red{of a SLAM algorithm} in an unseen environment $E'$. In particular, it predicts all the four components of $P_{E'}$ given $F_{E'}$. From it, we apply the feature extraction step of Section~\ref{S:featureextraction} to obtain $F_{E'}$. Note that $F_{E'}$ is calculated from the floor plan of $E'$, without requiring a robot to actually visit $E'$. We then use the models introduced in the previous section to find the components of $\hat{P}_{E'}$.

\begin{figure}[]
\centering
%\begin{subfloat}{
% \includegraphics[width=.4\linewidth]{pictures/cap7/simulations/freiburg52_GT}}
%\end{subfloat}%
\begin{subfloat}{
  \includegraphics[width=.35\linewidth]{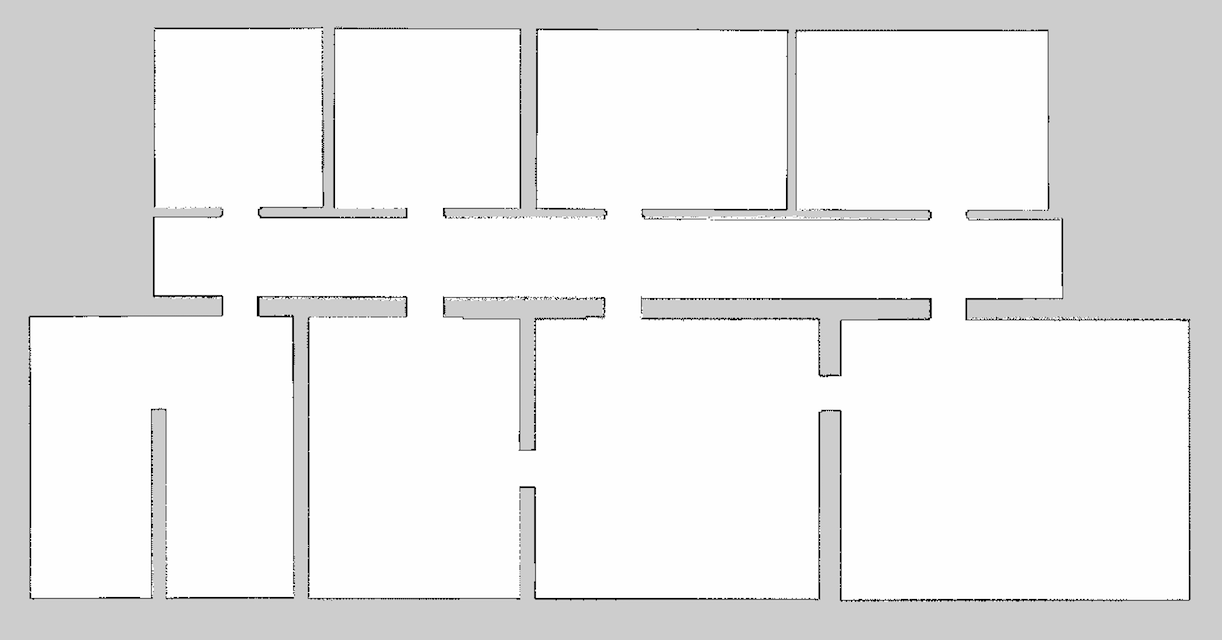}}
\end{subfloat}%\\
%\begin{subfloat}
%{
 % \includegraphics[width=.4\linewidth]{pictures/cap7/simulations/lab_b_GT}}
%\end{subfloat}%
\begin{subfloat}
{
  \includegraphics[width=.49\linewidth]{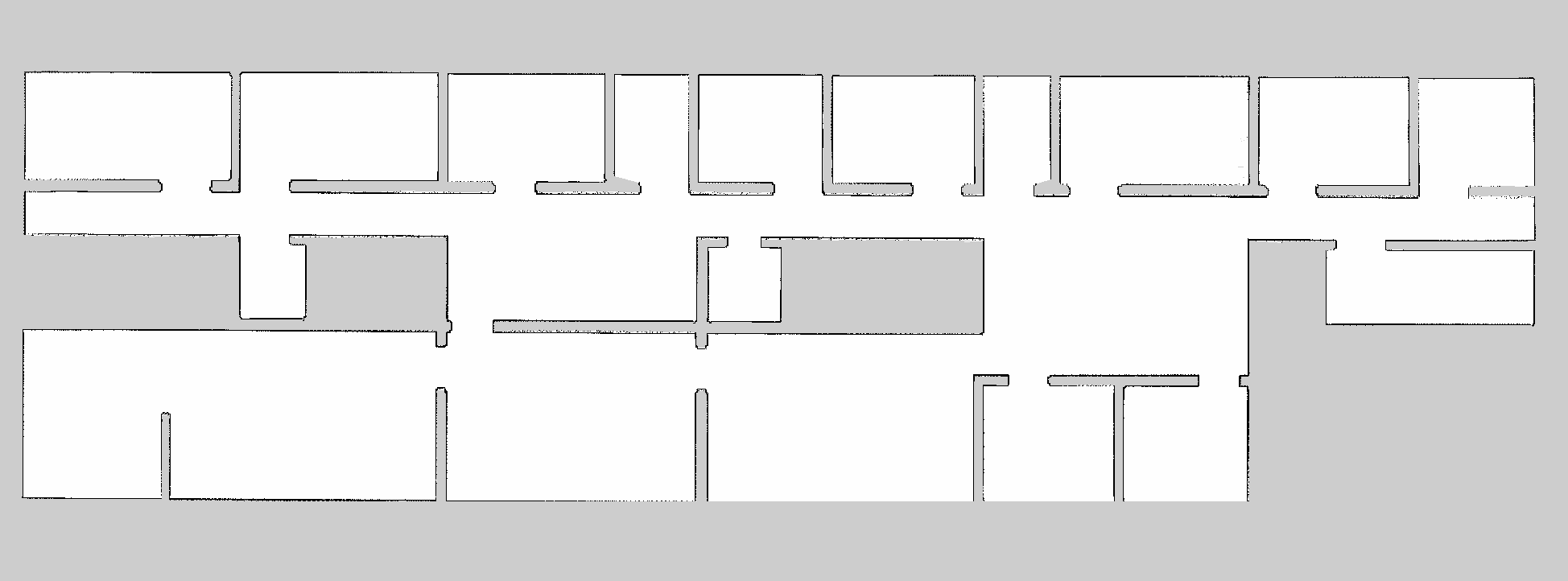}}
\end{subfloat}
\caption{The maps obtained in simulation of the Freiburg 52 (left) and Bremen (right) environments.}
\label{fig:simulation_test_freiburg}
\end{figure}

As an example, consider the environments in Fig.~\ref{fig:simulation_test_freiburg} from the dataset of~\cite{7487234} that were not included in the training set $\mathcal{E}$. Tables \ref{tab:simulation_generalization} and \ref{tab:simulation_generalization_2} show the comparison between the values predicted by our models and the localization error obtained in simulation. Our predictions are quite close the measured values of the localization error. 
%In particular, for the first building, the predicted error on the translational component is within $18$\% and $11$\% for the mean and the standard deviation, respectively, while the prediction of the mean of the rotational localization error is off by less than $8$\%. As expected, the standard deviation of the rotational component shows a less satisfactory prediction error of about $32$\%, which however amounts to an absolute difference of less than $\SI[per-mode = symbol]{0.05}{\degree}$.
%
%For the second environment, the prediction error on the translational component is below $13$\% and $2$\% for the mean and the standard deviation respectively, while our estimate of the mean of the rotational localization error is off by less than $4$\% with respect to the simulation data. As before, the standard deviation of the rotational component shows a higher prediction error of about $21$\%, which however amounts to an absolute error of less than $\SI[per-mode = symbol]{0.07}{\degree}$.

It is also interesting to notice that our method is able to predict the fact that the performance of \red{SLAM method considered here, GMapping,} in the Bremen environment  is worse than that in the Freiburg 52 environment. Note that this is not intuitive, since the two environments have a similar structure. This prediction is confirmed by the actual localization error measured in simulations, showing that the proposed method could be used to identify environments that are expected to be ``easy'' and ``hard'' for \red{a given SLAM method}.
%\textcolor{red}{abbiamo un esempio di questo tipo anche per Karto? o teniamo lo stesso?}
%\textcolor{blue}{INSERIRE CONSIDERAZIONE SUL FATTO CHE VALE PER ENTRAMBI I METODI QUANTO FATTO?}.

\begin{table}[]
\centering
\caption{Translational and rotational components of the localization error for the Freiburg 52 environment.}
\label{tab:simulation_generalization}
\begin{tabular}{l|c|c|c|c|}
\cline{2-5}
 & \multicolumn{2}{c|}{Translational error [m]} & \multicolumn{2}{c|}{Rotational error [rad]} \\ \cline{2-5} 
 & $\mean{\varepsilon_{t}()}$ & $s(\varepsilon_{t}())$ & $\mean{\varepsilon_{r}()}$ & $s(\varepsilon_{r}())$ \\ \hline
\multicolumn{1}{|l|}{Simulation} & 0.349 & 0.051 & 0.019 & 0.003 \\ \hline
\multicolumn{1}{|l|}{Prediction} & 0.286 & 0.046 & 0.020 & 0.003 \\ \hline
\end{tabular}
\end{table}

\begin{table}[]
\centering
\caption{Translational and rotational components of the localization error for the Bremen environment.}
\label{tab:simulation_generalization_2}
\begin{tabular}{l|c|c|c|c|}
\cline{2-5}
 & \multicolumn{2}{c|}{Translational error [m]} & \multicolumn{2}{c|}{Rotational error [rad]} \\ \cline{2-5} 
 & $\mean{\varepsilon_{t}()}$ & $s(\varepsilon_{t}())$ & $\mean{\varepsilon_{r}()}$ & $s(\varepsilon_{r}())$ \\ \hline
\multicolumn{1}{|l|}{Simulation} & 0.534 & 0.058 & 0.023 & 0.005 \\ \hline
\multicolumn{1}{|l|}{Prediction} & 0.469 & 0.057 & 0.024 & 0.004 \\ \hline
\end{tabular}
\end{table}

%The prediction module requires in input the floor plan of the environment whose localization error should be predicted, compute its VTD and VDR and selects the model with the lowest RMSE for the selected component.% Then, it loads all the models that have been computed by the data analysis module and that are available for prediction.
%
%The user is able to select which component of the localization error should be predicted and choose which prediction model should be used from a list of all the available models for the selected component. Alternatively, the user can specify only the component of the localization error that should be predicted and the prediction module automatically selects the model with the lowest RMSE for the selected component.
%
%The prediction module then proceeds to compute its prediction for the chosen component of the localization error.

%% file: 06-Exp.tex
\section{Experimental Validation}\label{sec:EXP}

In this section, we show the results of the validation of our method in different settings: on a publicly available dataset and on data collected by a real robot.

%A direct comparison between the predictions of our methodology and the actual localization errors obtained on some  publicly available dataset only provides limited information on the validity of our approach, since some features of the robots used to collect the datasets are different from those that we have considered to build our training set. This is especially true for sensor capabilities and accuracy of the odometry readings. ABBIAMO I DATI USATI PER LE NOSTRE SIMULAZIONI? SI SONO IN SECTION \ref{s:simulations}

We consider the dataset of \cite{sturm12iros}, composed of four real-robot runs inside an area of $10 \times 12$ $\SI[per-mode = symbol]{}{\meter\squared}$ of an L-shaped industrial hall, with associated ground truth data of the robot trajectories obtained with a motion capture system. We use two versions of the dataset: empty hall (without furniture) and with furniture. The characteristics of the laser range scanner onboard the robot and its systematic translational and rotational errors affecting the odometry are unfortunately not explicitly reported. From the analysis of the raw scans, we infer a field of view of $\SI[per-mode = symbol]{80}{\degree}$, an angular resolution of $\SI[per-mode = symbol]{1}{\degree}$, a range of about $\SI[per-mode = symbol]{30}{\meter}$, and a frequency of $\SI[per-mode = symbol]{10}{\hertz}$ for the laser range scanner and we make an educated guess assuming the translational and rotational odometry errors to be not greater than $\SI[per-mode = symbol]{0.01}{\meter\per\meter}$ and $\SI[per-mode = symbol]{2}{\degree\per\radian}$ respectively. 
%The remotely controlled robot is equipped with a front-facing laser range scanner, whose characteristics are unfortunately not explicitly reported in the documentation of the dataset; from the analysis of the raw scans, we infer a field of view of $\SI[per-mode = symbol]{80}{\degree}$, an angular resolution of $\SI[per-mode = symbol]{1}{\degree}$, a range of about $\SI[per-mode = symbol]{30}{\meter}$, and a frequency of $\SI[per-mode = symbol]{10}{\hertz}$.
%Similarly, the exact amount of systematic translational and rotational error affecting the odometry readings is not reported in the dataset's documentation. 
%Considering our prior experience with similar robots, we make an educated guess and assume the translational and rotational odometry error to be not greater than $\SI[per-mode = symbol]{0.01}{\meter\per\meter}$ and $\SI[per-mode = symbol]{2.0}{\degree\per\radian}$ respectively. 

The data recorded in the dataset are fed to \red{the SLAM method} and the reconstructed trajectory is compared to the ground truth trajectory using the performance metric of Section~\ref{S:metric} to obtain the localization error of the real robot (first row of Table~\ref{tab:dataset_results}). 
\red{For simplicity results discussed here are obtained using a single SLAM method, GMapping. However, as we have shown in Section \ref{S:featureextraction}, results can be generalized to other SLAM methods, as KartoSLAM.}

We perform $10$ simulations (as in Section \ref{S:simulations}) in the environments of~\cite{sturm12iros} using the same sensor and odometry configuration of the real robot and we report the average localization error in the second row of Table~\ref{tab:dataset_results}. We can say that our Stage simulations offer a rather accurate approximation of the behavior of a real robot. This result corroborates the validity of gathering training data for our model using simulations.

However, our training data ($P_{E}$ and $F_{E}$ for all $E \in \mathcal{E}$) have been obtained considering a different configuration for robot sensor and odometry (notably, considering a field of view of $\SI[per-mode = symbol]{270}{\degree}$). This different configuration has an impact on the accuracy of predictions, as we now show. Third row of Table~\ref{tab:dataset_results} shows the average localization error obtained from $10$ simulations performed with the robot configuration of Section \ref{S:simulations}. The fourth row of Table~\ref{tab:dataset_results} reports the localization error predicted by our method with the models of Section~\ref{s:modellearning}.
The difference between our predictions and the measured values of the localization error is due to different odometry accuracy and sensors' characteristics between the real robot used in \cite{sturm12iros} and the virtual robot employed to gather our training data. In particular, the limited field of view of the laser range scanner used by the real robot ($\SI[per-mode = symbol]{80}{\degree}$) negatively impacts the ability of GMapping to compensate for inaccuracies of the odometry, therefore leading to a higher localization error. The translational component $\varepsilon_{t}$ of the localization error is better predicted than the rotational component $\varepsilon_{r}$. This can be due to the fact that the estimation of the rotational odometry error used by our virtual robot disregards many aspects that can affect a real robot, like the state of the wheels and the slipperiness of the floor.

\begin{table}[]
\centering
\caption{Values of the translational and rotational components of the localization error, in m and rad respectively, for the dataset of \cite{sturm12iros}.}
\label{tab:dataset_results}
\begin{tabular}{l|l|l|l|l|}
\cline{2-5}
 & \multicolumn{2}{c|}{Empty hall} & \multicolumn{2}{c|}{Furniture} \\ \cline{2-5} 
 & \multicolumn{1}{c|}{$\varepsilon_{t}$} & \multicolumn{1}{c|}{$\varepsilon_{r}$} & \multicolumn{1}{c|}{$\varepsilon_{t}$} & \multicolumn{1}{c|}{$\varepsilon_{r}$} \\ \hline
\multicolumn{1}{|l|}{Dataset real robot} & 0.189 & 0.058 & 0.267 & 0.070 \\ \hline
\multicolumn{1}{|l|}{Simulated real robot} & 0.223 & 0.030 & 0.245 & 0.045 \\ \hline \hline
%\multicolumn{1}{|l|}{Optimistic simulation} & 0.073230 & 0.003982 & 0.089356 & 0.005462 \\ \hline
%\multicolumn{1}{|l|}{Optimistic prediction} & 0.083726 & 0.006582 & 0.087483 & 0.006640 \\ \hline
\multicolumn{1}{|l|}{Simulation virtual robot} & 0.146 & 0.023 & 0.164 & 0.024 \\ \hline
\multicolumn{1}{|l|}{Prediction virtual robot} & 0.128 & 0.018 & 0.138 & 0.018 \\ \hline
\end{tabular}
\end{table}

%\begin{figure}[htp]
%\centering
%\begin{subfloat}{
%  \includegraphics[width=0.25\linewidth]{pictures/cap7/dataset/robot_map.png}}
%\end{subfloat}%
%\begin{subfloat}{
%  \includegraphics[width=0.25\linewidth]{pictures/cap7/dataset/simulated_map.png}}
%\end{subfloat}%
%\caption{On the left, the map produced by GMapping using the recorded sensory information of the dataset. On the right, the map produced by GMapping on a simulation performed with the same robot setting used to gather the original dataset.}
%\label{fig:map_comparison_dataset}
%\end{figure}

Our method is able to provide more reliable predictions when the robot configuration is the same as that used for building the training set (third and fourth rows of Table~\ref{tab:dataset_results}). 
This is particularly true for the translational component $\varepsilon_{t}$ of the localization error.
%, where our predictions being off by less than $13$\% in the empty hall scenario and less than $16$\% in the furniture scenario.
%Instead, the results for the rotational component of the localization error show a slightly lower degree of accuracy, with our predictions being off by about $30$\% on average.

\begin{figure}[]
\centering
\begin{subfloat}{
  \centering
  \includegraphics[width=0.283\linewidth]{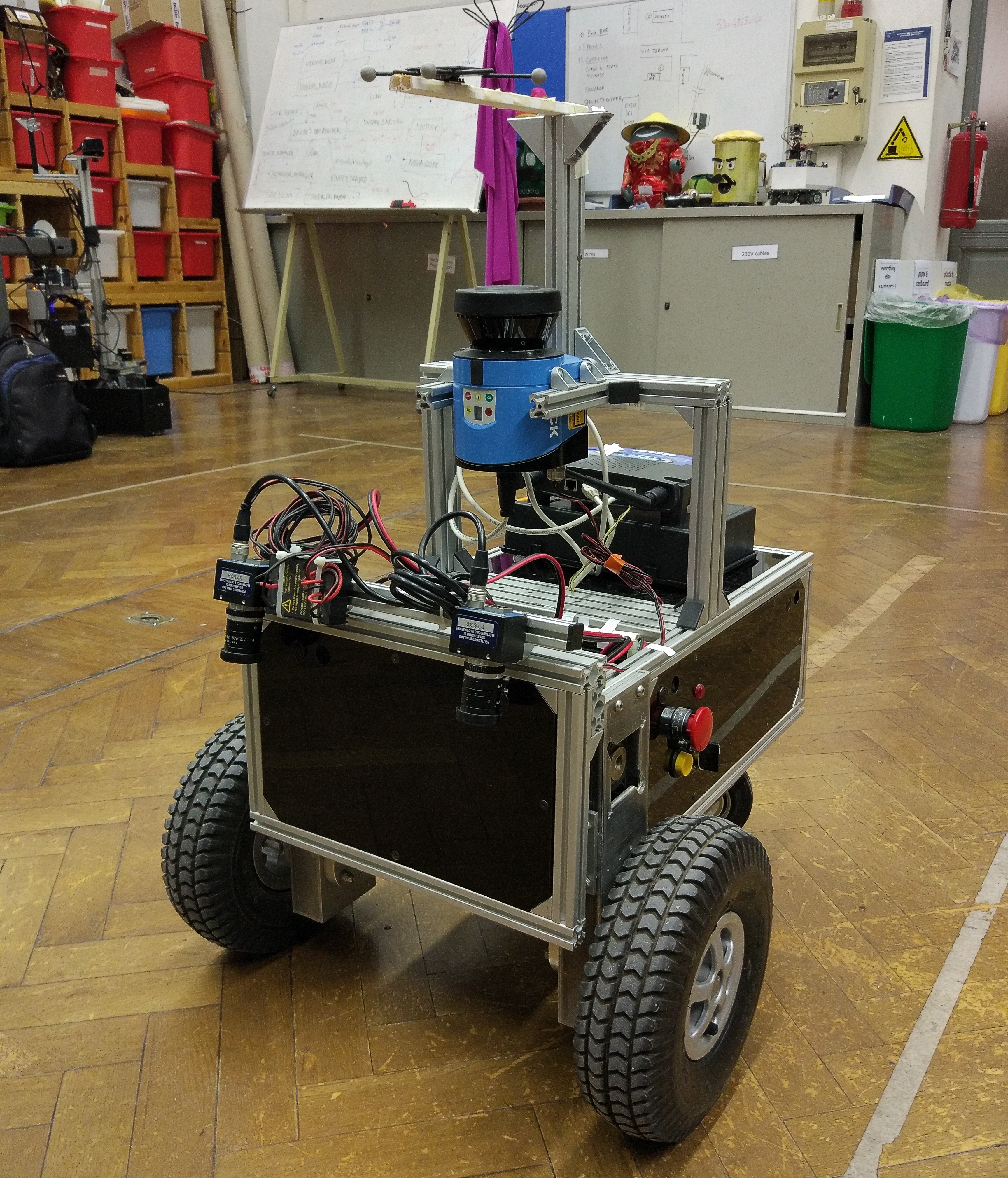}}
\end{subfloat}%
\begin{subfloat}{
  \centering
  \includegraphics[width=0.33\linewidth]{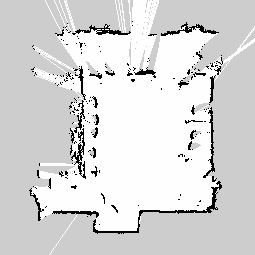}}
\end{subfloat}%
\begin{subfloat}{
  \includegraphics[width=0.33\linewidth]{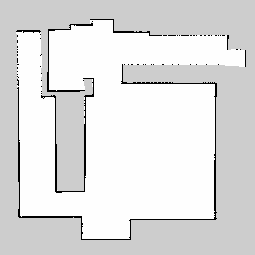}}
\end{subfloat}
\caption{The robot Robocom (left). The map built by Robocom in the AIRLab (center). The map build in a simulation run (right).}
\label{fig:map_comparison_robocom}
\end{figure}

As further validation of our method, we collect data in the Artificial Intelligence and Robotics Laboratory (AIRLab) at the Politecnico di Milano. 
The environment consists of a square $9 \times 9$ $\SI[per-mode = symbol]{}{\meter\squared}$ hall (see Fig.~\ref{fig:map_comparison_robocom} right).
%, approximately divided in two zones: an L-shaped area hosting tables, desks and shelves with experimental materials, and an empty square $6 \times 6$ $\SI[per-mode = symbol]{}{\meter\squared}$ area.
We employ a three-wheeled differential drive robot, called Robocom, 
equipped with with a SICK LMS100 laser range scanner, with a field of view of $\SI[per-mode = symbol]{270}{\degree}$, an angular resolution of up to $\SI[per-mode = symbol]{0.25}{\degree}$, a range of $\SI[per-mode = symbol]{20}{\meter}$, and a frequency of $\SI[per-mode = symbol]{50}{\hertz}$ (Fig.~\ref{fig:map_comparison_robocom} left). We manually estimate the amount of systematic translational and rotational errors affecting the odometry as to be not greater than $\SI[per-mode = symbol]{0.01}{\meter\per\meter}$ and $\SI[per-mode = symbol]{4}{\degree\per\radian}$, respectively. Note that the above configuration is similar, but not identical, to that of the virtual robot of Section~\ref{S:simulations} used to collect our training data. In this way, we aim at showing the robustness of our method that can predict performance also for non-identically configured robots.
We perform $10$ runs, each involving the autonomous exploration of the area (Fig.~\ref{fig:map_comparison_robocom} center), which is covered by an OptiTrack motion capture system to collect ground truth trajectory. Explorations are performed according to the same methodology we described in Section \ref{S:simulations}. We also perform $10$ simulations, as before, using the Robocom configuration (Fig.~\ref{fig:map_comparison_robocom} right).

%
%\begin{figure}[htp]
%\centering
%  \includegraphics[width=0.4\linewidth]{pictures/cap7/robocom/robocom_2_smaller.jpg}
%\caption{The Robocom robotic platform.}
%\label{fig:robocom}
%\end{figure}

\begin{table}[]
\centering
\caption{Values of the translational and rotational components of the localization error for the AIRLab experiment.}
\label{tab:robocom_results}
\begin{tabular}{l|l|l|l|l|}
\cline{2-5}
 & \multicolumn{2}{c|}{Translational error [m]} & \multicolumn{2}{c|}{Rotational error [rad]} \\ \cline{2-5} 
 & \multicolumn{1}{c|}{$\mean{\varepsilon_{t}()}$} & \multicolumn{1}{c|}{$s(\varepsilon_{t}())$} & \multicolumn{1}{c|}{$\mean{\varepsilon_{r}()}$} & \multicolumn{1}{c|}{$s(\varepsilon_{r}())$} \\ \hline
\multicolumn{1}{|l|}{Robocom} & 0.086 & 0.026 & 0.066 & 0.010 \\ \hline
\multicolumn{1}{|l|}{Simulation} & 0.101 & 0.019 & 0.022 & 0.004 \\ \hline
\multicolumn{1}{|l|}{Prediction} & 0.111 & 0.032 & 0.018 & 0.003 \\ \hline
\end{tabular}
\end{table}

The results of the validation are shown in Table \ref{tab:robocom_results}. A first consideration is that the difference between $\mean{\varepsilon_{t}()}$ of the simulations and that of the real robot is small (a difference of $\SI[per-mode = symbol]{2.5}{\centi\meter}$ on the mean and less than $\SI[per-mode = symbol]{1}{\centi\meter}$ on the standard deviation). 
The predicted value of $\mean{\varepsilon_{r}()}$ is less consistent. We believe this discrepancy to be mostly due to the limitations of our own Robocom robotic platform. 
%While the estimated $\SI[per-mode = symbol]{0.01}{\meter\per\meter}$ translational odometry error of Robocom coincides with the value of the simulation setting and represents a realistic estimate of the performance of a generic wheeled robot, 
The estimated $\SI[per-mode = symbol]{4}{\degree\per\radian}$ rotational odometry error is effectively twice as high as the value of the virtual robot configuration and is probably close to the maximum rotational error that can be tolerated while still achieving acceptable SLAM performance. 

Overall, the validation shows that our method is very effective in predicting the localization error of \red{SLAM method} in unseen environments when the robot platform is identical to that used to collect training data (Tables~\ref{tab:simulation_generalization} and~\ref{tab:simulation_generalization_2}). When this does not hold, our method is still able to provide very good predictions. For example, in Table~\ref{tab:robocom_results}, a prediction of the translational error that is wrong of $\SI[per-mode = symbol]{2.5}{\centi\meter}$, and that is obtained without running the SLAM algorithm and sending the robot in the environment, can be considered quite satisfactory for most applications.
\blue{However, if the robotic platform used for a target application presents significant differences (e.g. a laser range of $\SI[per-mode = symbol]{3.5}{\meter}$ against one of $\SI[per-mode = symbol]{30}{\meter}$, $\SI[per-mode = symbol]{180}{\degree}$ or $\SI[per-mode = symbol]{360}{\degree}$ of FOV), the collection of a new training dataset is required. This training phase could be performed automatically by updating the parameters in the simulator. Multiple runs in multiple environments are automatically executed in batch by our proposed method, without the need for a supervision.} 

%As for the validation of our approach, the results show that the predictions of our models are substantially consistent with the localization error of the simulations. 
%In particular, the predicted means of the translational and rotational components of the localization error are remarkably close to the corresponding simulation values (a difference of about $\SI[per-mode = symbol]{1}{\centi\meter}$ on the translational component and less than $\SI[per-mode = symbol]{0.3}{\degree}$ on the rotational component). The prediction error committed on the standard deviation of the rotational localization error is even smaller, with the prediction being off by about $9$\%. 

%The most significant exception is represented by the standard deviation of the translational localization error, which in this case is closer to the one committed by the real robot than the one measured on the simulation runs. However, in absolute terms the prediction error is still smaller than $\SI[per-mode = symbol]{1.4}{\centi\meter}$.

%Finally, if we compare the predicted mean of the translational localization error with its corresponding value measured during runs of real robot, we can see that the prediction is overestimating the true error by $\SI[per-mode = symbol]{2.5}{\centi\meter}$, which is more than twice with respect to the simulations, but still quite good for predictively assessing the performance of GMapping in the environment.

%COMMENTO SU VALIDITA': SBAGLIARE DI POCO MA SENZA ENTRARE NELL'AMBIENTE.

%% file: 07-Con.tex
\section{Conclusions}\label{sec:CON}
In this paper, we presented a method for predictive benchmarking of SLAM algorithms. Our method collects the performance of a SLAM algorithm in several environments, builds a model of the relationship between the performance and the features of the environments, and uses this model to predict the performance of the SLAM algorithm given the features of an unseen environment. Validation of our method on data from publicly available datasets and collected with real robots shows that it can actually be employed to predict the performance of GMapping in unseen environments.
\red{Our method is based on linear regression, which constitutes a positive feature of our approach as, due to its interpretability, facilitates its use in different settings.}

This paper has not addressed the applications of the proposed predictive benchmarking method. Future work will apply the proposed method to other SLAM algorithms in order to enable the full potential of the approach for selecting the most appropriate SLAM algorithm at design time, before the robot is actually deployed in a new environment. Moreover, the integration with traditional \emph{ex-post} benchmarking tools will be investigated.

\red{Future works involve the extension of the proposed framework for benchmarking other robot-related tasks and the assessment of 3D SLAM performances. Note that, in order to generalize our method to other domains, two steps are needed: the selection of an evaluation metric, and the fully-automated collections of runs in the dataset. This second step, however, could be avoided for tasks that could be performed using a 2D dataset, due to the availability of the dataset of \cite{OURIROS} we used here. Tasks involving a 3D perception of the environment, like 3D SLAM, could exploit the use of publicly available datasets like \cite{gibson}. Finally, note that the extension of our proposed method to 3D SLAM will require only the second step, as the metric of \cite{Kuemmerle2009} could be applied also to 3D settings.}